%% file: main.tex
\documentclass{article}

\usepackage[nonatbib, final]{neurips_2021}
\usepackage[numbers,sort&compress]{natbib}

\usepackage[utf8]{inputenc} % allow utf-8 input
\usepackage[T1]{fontenc}    % use 8-bit T1 fonts
\usepackage{hyperref}       % hyperlinks
\usepackage{url}            % simple URL typesetting
\usepackage{booktabs}       % professional-quality tables
\usepackage{amsfonts}       % blackboard math symbols
\usepackage{nicefrac}       % compact symbols for 1/2, etc.
\usepackage{microtype}      % microtypography
\usepackage{amsthm}
\usepackage{algorithm}
\usepackage{algpseudocode}
\usepackage{mathtools}
\usepackage{graphicx}
\usepackage{subcaption}
\usepackage{tabularx}
\usepackage{dsfont}
\usepackage{multirow}
\usepackage{wrapfig}

\usepackage{color}

\usepackage{multicol}

\usepackage{caption}

\usepackage{enumitem}
\usepackage[dvipsnames]{xcolor}

 \usepackage{tikz}
    \usetikzlibrary{trees, shapes,arrows,automata,positioning,cd}
    \tikzset{
      cfgedge/.style   = {black, ->, >=stealth},
      forward/.style = { blue, ->, >=angle 45},
      backward/.style = { red, densely dashed, ->, >=latex' },
      backwardleft/.style = { red, densely dashed, <-, >=latex' },
    }
    
\DeclareMathOperator*{\argmin}{\arg\!\min}

\theoremstyle{definition}
\newtheorem{definition}{Definition}[section]

\newtheorem{theorem}{Theorem}[section]

\title{Fair Clustering Using Antidote Data}

\author{%
  Anshuman Chhabra\textsuperscript{*}, Adish Singla\textsuperscript{\textdagger}, Prasant Mohapatra\textsuperscript{*}\\
  \textsuperscript{*} University of California, Davis, \texttt{\{chhabra, pmohapatra\}@ucdavis.edu}\\
   \textsuperscript{\textdagger} MPI-SWS, \texttt{adishs@mpi-sws.org}\\
}

\begin{document}

\captionsetup[table]{font=footnotesize, skip=0pt}
\captionsetup[figure]{font=footnotesize, skip=0pt}

\maketitle

\begin{abstract}
Clustering algorithms are widely utilized for many modern data science applications. This motivates the need to make outputs of clustering algorithms fair. Traditionally, new \textit{fair} algorithmic variants to clustering algorithms are developed for specific notions of fairness. However, depending on the application context, different definitions of fairness might need to be employed. As a result, new algorithms and analysis need to be proposed for each combination of clustering algorithm and fairness definition. Additionally, each new algorithm would need to be reimplemented for deployment in a real-world system. Hence, we propose an alternate approach to \textit{group-level} fairness in \textit{center-based} clustering inspired by research on \textit{data poisoning attacks}. We seek to augment the original dataset with a small number of data points, called \textit{antidote} data. When clustering is undertaken on this new dataset, the output is \textit{fair}, for the chosen clustering algorithm and fairness definition. We formulate this as a general bi-level optimization problem which can accommodate any center-based clustering algorithms and fairness notions. We then categorize approaches for solving this bi-level optimization for two different problem settings. Extensive experiments on different clustering algorithms and fairness notions show that our algorithms can achieve desired levels of fairness on many real-world datasets with a very small percentage of antidote data added. We also find that our algorithms achieve lower fairness costs and competitive clustering performance compared to other state-of-the-art fair clustering algorithms.
\end{abstract}

\section{Introduction}
\input{introduction}

\section{Problem Statement}
\input{problem_statement}

\section{Proposed Approaches}
\input{proposed_approaches}

\section{Results}

\input{results}

\section{Related Works}
\input{related_works}

%\section{Limitations}
%\input{discussion}

\section{Concluding Discussions}
\input{conclusion}

\bibliographystyle{unsrt}

\input{main.bbl}
%\bibliography{dblp_ref.bib}
%% For arXiv, directly paste bbl and comment above line

%%%%%%%%%%%%%%%%%%%%%%%%%%%%%%%%%%%%%%%%%%%%%%%%%%%%%%%%%%%%

\end{document}

\captionsetup[table]{font=footnotesize, skip=0pt}
\captionsetup[figure]{font=footnotesize, skip=0pt}

\maketitle

\tableofcontents

\newpage

\section{Proofs and Derivations for Results in the Main Paper}

\subsection{Proof for Theorem 3.1}

\begin{manualtheorem}{3.1}\label{other_paper}\citep{yu2014sampling}.
\textit{Let $V^* \in \mathbb{R}^{V_s^{(t)} \times d}$ be a minimizer for the function $f(V)$ in an iteration $t$ of Algorithm 3 and for $\epsilon > 0$ define $X = \{V \in \mathbb{R}^{V_s^{(t)} \times d} \, | \, f(V) - f(V^*) \leq \epsilon\}$. Let $\mathbb{P}_{X}$ denote the average probability of successfully sampling from the uniform distribution over $X$ by algorithm $\mathcal{A}$, and it takes $n_{X}$ samples to realize $\mathbb{P}_{X}$. Then, the number of queries to $f$ that $\mathcal{A}$ makes to compute $\tilde{V}$ s.t. $f(\tilde{V}) - f(V^*) \leq \epsilon$  with probability at least $1 - \delta$ is bounded as $\mathcal{O}(\max\{\frac{\ln{(\delta^{-1})}}{\mathbb{P}_{X}}, n_{X}\})$.}
\end{manualtheorem}

\begin{proof}
The proof follows from Theorem 1 in \citep{yu2014sampling}. Since we are considering a sampling-only framework, we set $\lambda = 1$ in Theorem 1 \citep{yu2014sampling} to obtain the result.
\end{proof}

\subsection{Derivations for Section 3.1}
In this subsection we discuss the derivation of the single-level reduction using KKT constraints with $\mathcal{C}_{\textrm{SON}}$ and $\mathcal{F}_{\textrm{social}}$. First, consider the original strongly convex SON clustering objective:
\begin{equation}
    \min_{\mu' \in \mathbb{R}^{n \times d}} \frac{1}{2}\sum_{j=1}^n||U_j - \mu_j'||^2 + \lambda\sum_{i<j}||\mu_i' - \mu_j'||
\end{equation}

As described in the main text, we create the ordering $O$, the graph $G$ and define its node-arc-incidence matrix $I$ \citep{node-arc} and then reformulate the above objective:

\begin{equation}
    \min_{\mu \, \in \, \mathbb{R}^{n \times d}, \; \eta \, \in \, \mathbb{R}^{d \times |O|}} \;\; \frac{1}{2} ||\mu - U||^2 + \lambda \sum_{i \in O}||\eta^i|| \;\;\textrm{ s.t.} \;\;\,\, \mu^{T}I - \eta = 0
\end{equation}

It can be verified that objectives (5) and (6) are equivalent. We can even define the dual formulation for the above primal problem (where $\langle , \rangle$ denotes the matrix Frobenius inner-product):
\begin{equation}
    \begin{aligned}
    \max_{\theta \in \mathbb{R}^{n \times d}, \, \zeta \in \mathbb{R}^{d \times |O|}} \quad &  \langle U^T, \theta \rangle - \frac{1}{2}||\theta||^2\\
\textrm{s.t.} \quad &  I\zeta^T - \theta = 0\\
& ||\zeta_i|| \leq \lambda, \forall i \in O\\
    \end{aligned}
\end{equation}

Now, we discuss the KKT conditions. Since the SON objective is strongly convex, we can use the reformulated primal (6) and dual (7) problems to arrive at the KKT conditions: 

\begin{equation}
    \begin{aligned}
    & \theta + \mu - U = 0\\
    & \eta - \mathcal{P}(\eta + \zeta) = 0\\
    & \mu^{T}I - \eta = 0\\
     & I\zeta^T - \theta = 0\\
\end{aligned}
\end{equation}

Here $\mathcal{P}(.)$ refers to the proximal operator of the Euclidean norm, therefore $\mathcal{P}(\eta + \zeta) = \max\{0, 1-\frac{1}{||\eta + \zeta||}\}(\eta + \zeta)$. Since we now have the KKT conditions we can undertake the single-level reduction for problem P1.R. 

For this we first have to substitute $U$ for $U\cup V$. As in the main text, $V_s^{(t)}$ denotes $|V|$ in iteration $t$ of Algorithm 1. The number of centers we have will thus be $\mu \in \mathbb{R}^{m \times d}$ where $m = n + V_s^{(t)}$ for $U \cup V$. So now we can use the KKT conditions by replacing $U$ with $U\cup V$ and $n$ with $m$. The other variables will also then be: $\mu \in \mathbb{R}^{m \times d}, \eta \in \mathbb{R}^{d \times |O|}, \theta \in \mathbb{R}^{m \times d}, \zeta \in \mathbb{R}^{d \times |O|}$. The original problem P1.R for $\mathcal{C}_{\textrm{SON}}$ and $\mathcal{F}_{\textrm{social}}$ is:

\begin{equation}
 \begin{aligned}
\min_{V, \, \mu} \quad & \mathcal{F}_{\textrm{social}}(\mu, U) \\
\textrm{s.t.} \quad & \mu = \mathcal{C}_{\textrm{SON}}(U \, \textstyle \cup \, V)\\
\end{aligned}   
\end{equation}

Replacing (8) as constraints for the upper-level objective in (9) and removing the lower-level objective gives us the single-level optimization problem we present in the main paper:

\begin{equation*}
 \begin{aligned}
\min_{V, \, \mu, \, \eta, \, \theta, \, \zeta } \quad & \mathcal{F}_{\textrm{social}}(\mu, U) \\
\textrm{s.t.} \quad & \theta + \mu - (U\cup V) = 0\\
     & \eta - \max\{0, 1-\frac{1}{||\eta + \zeta||}\}(\eta + \zeta) = 0\\
     & \mu^{T}I - \eta = 0\\
     & I\zeta^T - \theta = 0\\
\end{aligned}   
\end{equation*}

%\newpage

%\newpage
\section{Experiments for Algorithm 1 with $\mathcal{C}_{\textrm{SON}}$ and $\mathcal{F}_{\textrm{social}}$}

We provide results for Algorithm 1 when using CVX as the solver \citep{cvxpy} for the KKT reformulated single-level objective with fairness cost as $\mathcal{F}_{\textrm{social}}$. We show how the antidote data computed by our optimization consistently reduces the fairness cost compared to vanilla SON clustering while varying the regularization parameter $\lambda$ from 0.001 to 0.01. That is, we are showcasing the trend in fairness (cost) as a function of the \textit{number of clusters} (ie, number of unique centers) which are determined by the value of $\lambda$. 

For a given $\lambda$ we run Algorithm 1 and obtain $V$, and then re-run regular SON clustering on $U \cup V$. We then compare this obtained fairness cost to the one obtained by regular vanilla SON clustering on $U$. Note that since CVX does not scale well with large inputs, we subsample all datasets to 100. We also let $\gamma = 0.99$ and for all experiments we obtain $|V| \leq 10$, ie $|V|/|U| \leq 0.1$. We now present the results in Figure 1. 

Each of the 4 figures corresponds to the 4 real-world datasets we consider. To further simplify the model we only consider 2 protected groups. As the vanilla SON fairness cost values and fairness cost values for the $V$ obtained by Algorithm 1 could vary widely due to $\lambda$, it would be hard to decipher their curves individually. Thus, instead, we present the results as a difference between the vanilla SON fairness cost $\mathcal{F}_{\textrm{social}}(\mu^{\textrm{vanilla}},U)$ and the fairness cost values we obtain as a result of Algorithm 1 denoted by $\mathcal{F}_{\textrm{social}}(\mu,U)$. That is, the y-axis of the figures represents $\mathcal{F}_{\textrm{social}}(\mu^{\textrm{vanilla}},U) - \mathcal{F}_{\textrm{social}}(\mu,U)$ and the x-axis represents $\lambda$. It is clear to see then that in each figure, if the difference curves are positive on the y-axis, Algorithm 1 outperforms vanilla SON. As can be seen, the centers obtained as a result of clustering on $U\cup V$ where $V$ is obtained from Algorithm 1, lead to more fair clusters than traditional SON clustering (for the $\mathcal{F}_{\textrm{social}}$ metric).

\begin{figure}[ht!]
\centering
\begin{subfigure}{.49\textwidth}
  \centering
    \includegraphics[scale=0.4]{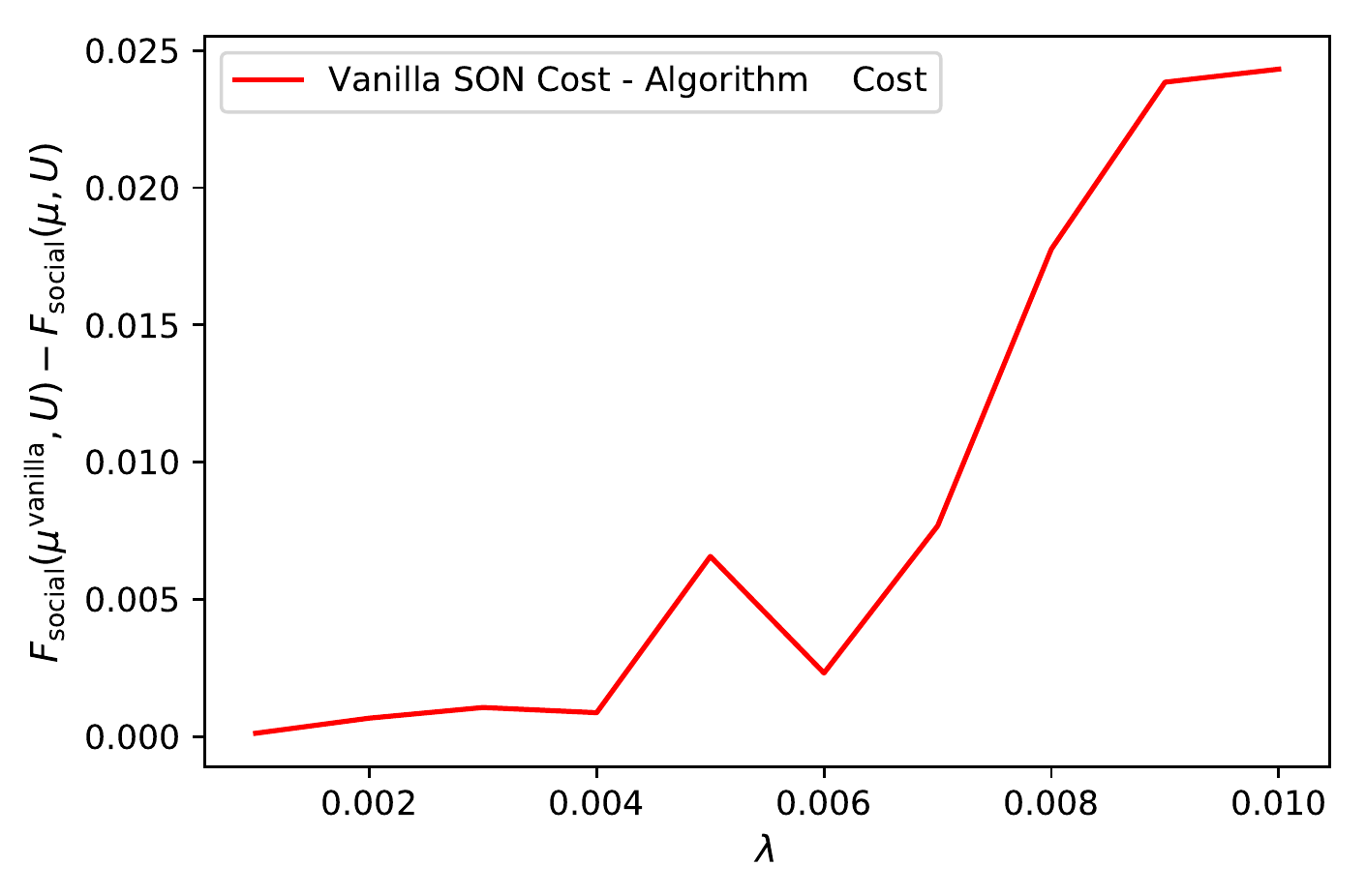}%
  \caption{Results for \texttt{adult}}
\end{subfigure}%
\begin{subfigure}{.49\textwidth}
  \centering
    \includegraphics[scale=0.4]{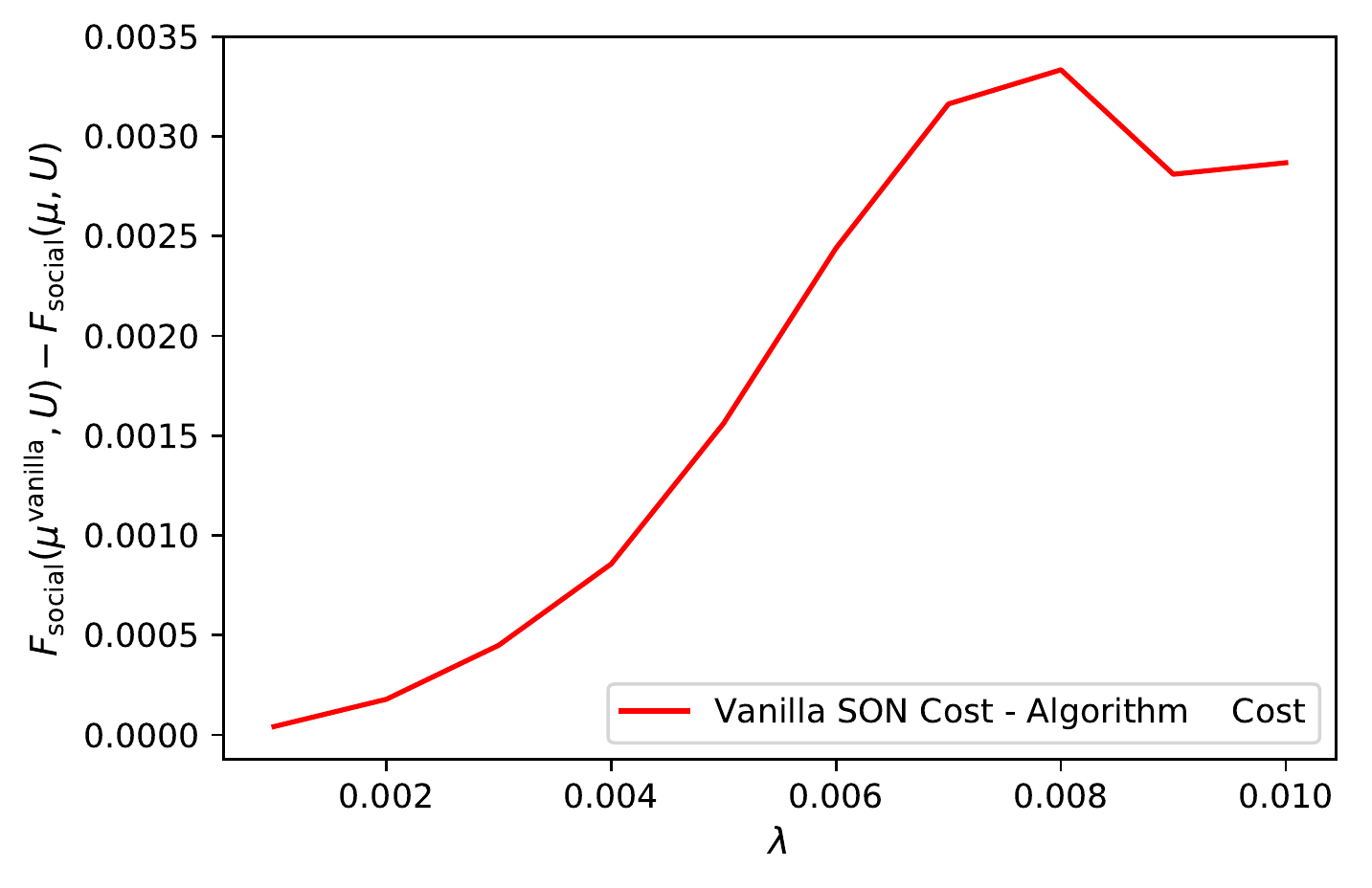}%
  \caption{Results for \texttt{bank}}
\end{subfigure}
\begin{subfigure}{.49\textwidth}
  \centering
    \includegraphics[scale=0.4]{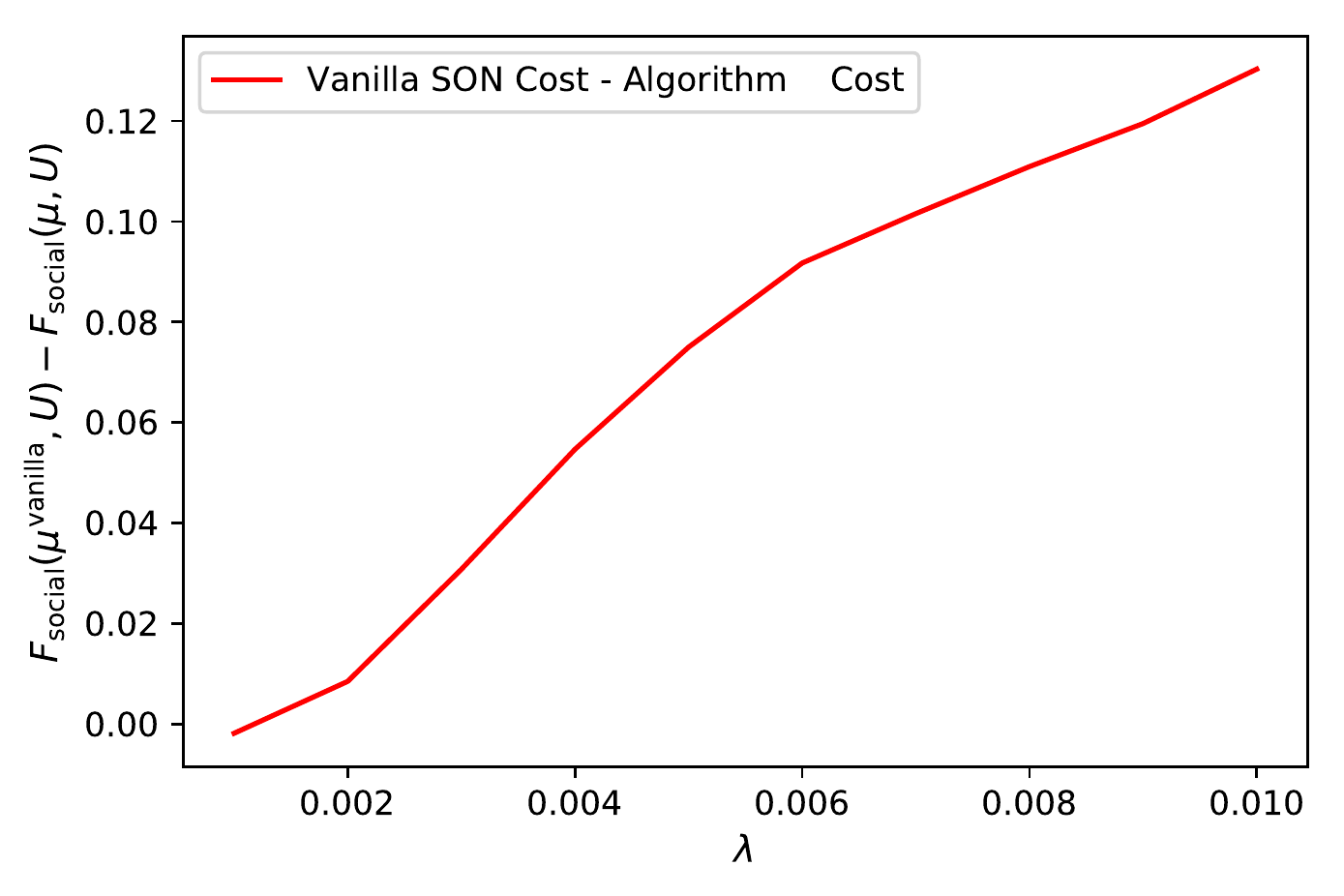}%
  \caption{Results for \texttt{creditcard}}
\end{subfigure}
\begin{subfigure}{.49\textwidth}
  \centering
    \includegraphics[scale=0.4]{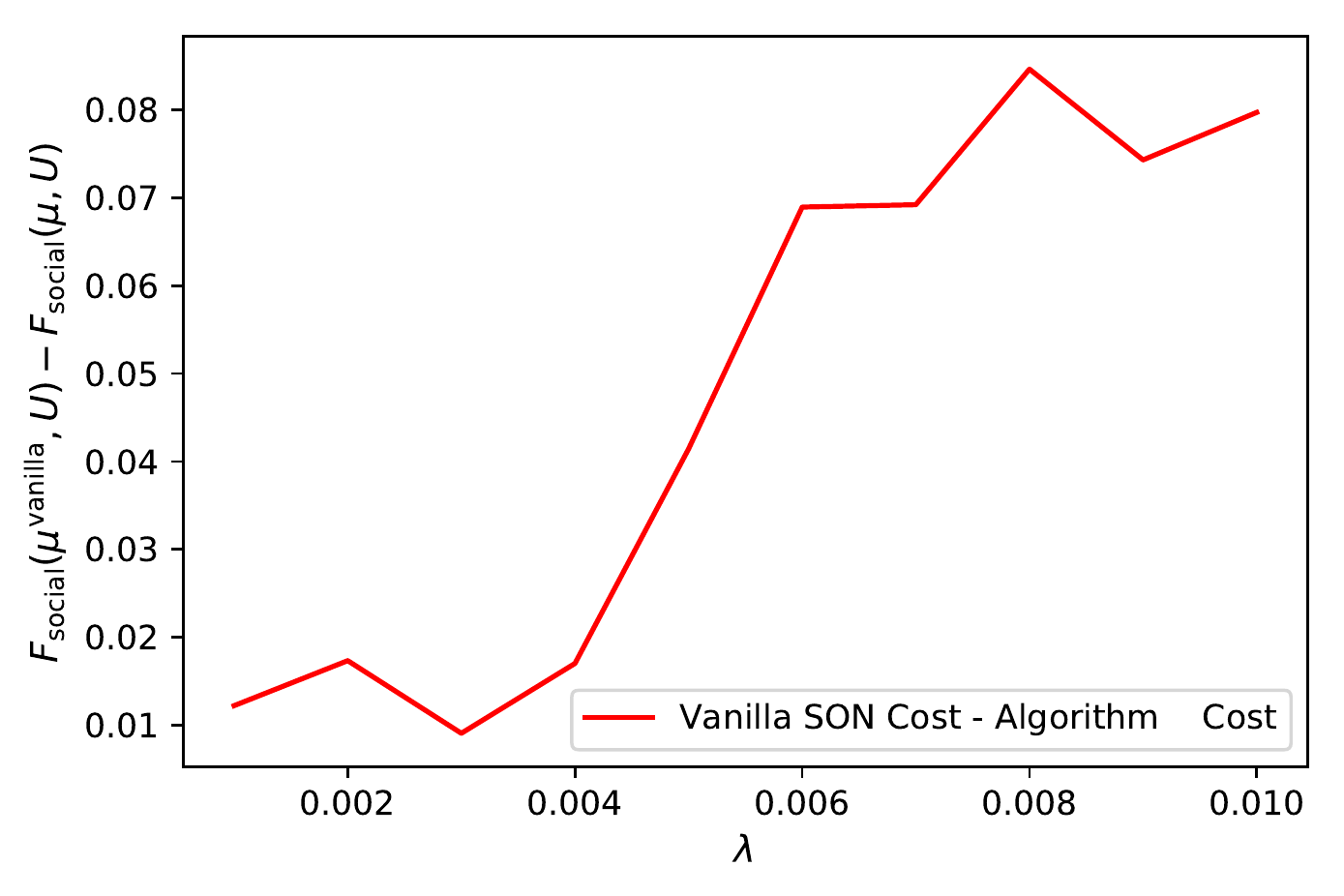}%
  \caption{Results for \texttt{LFW}}
\end{subfigure}
\caption{Results for Algorithm 1} 
\end{figure}

%\newpage
\section{Additional Experiments for Algorithm 2}

\subsection{Experiments for Algorithm 2 for $k=3$ and $k=4$}

As mentioned in the paper, we provide experiments for Algorithm 2 for $k$ greater than 2. We obtain results for $k=3$ and $k=4$ and present them in Table 1 and Table 2, respectively. The results are tabulated similar to the main paper-- we are comparing Algorithm 2 with vanilla clustering for each of the three combinations considered in the paper. As can be observed in both Table 1 and Table 2, Algorithm 2 can achieve improved fairness than traditional clustering. Another interesting trend that can be noted is that more antidote data needs to be added for larger $k$, in general. This can be seen by observing the $|V|/|U|$ values for both tables individually. This trend is intuitive however, as having more number of clusters can introduce more complexities in reducing fairness cost thus requiring more antidote data addition.

\begin{table}[ht!]
\fontsize{8}{8}\selectfont
\caption{Results for Algorithm 2 when $k=3$}
\centering
\begin{tabular}{cc||cc||cc}
\hline
Clustering-Fairness Combination & Dataset & $\alpha$ & $|V|/|U|$ & $\mathcal{F}(\mu^{\textrm{vanilla}}, U)$ & $\mathcal{F}(\mu, U)$ \\ \hline
\multirow{4}{*}{\begin{tabular}[c]{@{}c@{}}Combination \#1:\\ $\mathcal{C}_{\textrm{k-means}}, \mathcal{F}_{\textrm{balance}}$\end{tabular}} & \texttt{adult} & 0.0 & 0.002 & 0.0 & \textbf{-0.6249} \\ %\cline{2-6} 
 & \texttt{bank} & -0.2919 & 0.00044 & -0.2919 & \textbf{-0.2945} \\ %\cline{2-6} 
 & \texttt{creditcard} & -0.7957 & 0.00067 & -0.7957 & \textbf{-0.7977} \\ %\cline{2-6} 
 & \texttt{LFW} & -0.7834 & 0.0015 & -0.7834 & \textbf{-0.7871} \\ \hline
\multirow{4}{*}{\begin{tabular}[c]{@{}c@{}}Combination \#2:\\ $\mathcal{C}_{\textrm{k-means}}, \mathcal{F}_{\textrm{social}}$\end{tabular}} & \texttt{adult} & 3.4898 & 0.002 & 3.4898 & \textbf{3.4848} \\ %\cline{2-6} 
 & \texttt{bank} & 1.7985 & 0.00044 & 1.7985 & \textbf{1.7983} \\ %\cline{2-6} 
 & \texttt{creditcard} & 17.177 & 0.00067 & 17.177 & \textbf{17.165} \\ %\cline{2-6} 
 & \texttt{LFW} & 1308.099 & 0.0015 & 1308.099 & \textbf{1307.841} \\ \hline
\multirow{4}{*}{\begin{tabular}[c]{@{}c@{}}Combination \#3:\\ $\mathcal{C}_{\textrm{spectral}}, \mathcal{F}_{\textrm{balance}}$\end{tabular}} & \texttt{adult} & -0.543 & 0.02 & -0.543 & \textbf{-0.576} \\ %\cline{2-6} 
 & \texttt{bank} & -0.4216 & 0.02 & -0.4216 & \textbf{-0.4298} \\ %\cline{2-6} 
 & \texttt{creditcard} & -0.8056 & 0.02 & -0.8056 & \textbf{-0.8261} \\ %\cline{2-6} 
 & \texttt{LFW} & -0.7736 & 0.02 & -0.7736 & \textbf{-0.7938} \\ \hline
\end{tabular}
\end{table}

\begin{table}[ht!]
\fontsize{8}{8}\selectfont
\caption{Results for Algorithm 2 when $k=4$}
\centering
\begin{tabular}{cc||cc||cc}
\hline
Clustering-Fairness Combination & Dataset & $\alpha$ & $|V|/|U|$ & $\mathcal{F}(\mu^{\textrm{vanilla}}, U)$ & $\mathcal{F}(\mu, U)$ \\ \hline
\multirow{4}{*}{\begin{tabular}[c]{@{}c@{}}Combination \#1:\\ $\mathcal{C}_{\textrm{k-means}}, \mathcal{F}_{\textrm{balance}}$\end{tabular}} & \texttt{adult} & 0.0 & 0.15 & 0.0 & \textbf{-0.5384} \\ %\cline{2-6} 
 & \texttt{bank} & -0.2821 & 0.0011 & -0.2821 & \textbf{-0.2837} \\ %\cline{2-6} 
 & \texttt{creditcard} & -0.7694 & 0.0017 & -0.7694 & \textbf{-0.7705} \\ %\cline{2-6} 
 & \texttt{LFW} & -0.7383 & 0.0038 & -0.7383 & \textbf{-0.7691} \\ \hline
\multirow{4}{*}{\begin{tabular}[c]{@{}c@{}}Combination \#2:\\ $\mathcal{C}_{\textrm{k-means}}, \mathcal{F}_{\textrm{social}}$\end{tabular}} & \texttt{adult} & 3.0299 & 0.005 & 3.0299 & \textbf{3.01} \\ %\cline{2-6} 
 & \texttt{bank} & 1.315 & 0.0011 & 1.315 & \textbf{1.301} \\ %\cline{2-6} 
 & \texttt{creditcard} & 15.989 & 0.0017 & 15.989 & \textbf{15.976} \\ %\cline{2-6} 
 & \texttt{LFW} & 1242.913 & 0.0038 & 1242.913 & \textbf{1242.570} \\ \hline
\multirow{4}{*}{\begin{tabular}[c]{@{}c@{}}Combination \#3:\\ $\mathcal{C}_{\textrm{spectral}}, \mathcal{F}_{\textrm{balance}}$\end{tabular}} & \texttt{adult} & -0.3584 & 0.05 & -0.3584 & \textbf{-0.5354} \\ %\cline{2-6} 
 & \texttt{bank} & -0.273 & 0.05 & -0.273 & \textbf{-0.476} \\ %\cline{2-6} 
 & \texttt{creditcard} & -0.8425 & 0.05 & -0.8425 & \textbf{-0.9307} \\ %\cline{2-6} 
 & \texttt{LFW} & -0.7574 & 0.065 & -0.7574 & \textbf{-0.7658} \\ \hline
\end{tabular}
\end{table}

\subsection{Experiments Comparing Clustering Performance on Davies-Bouldin index \citep{dbindex} and Calinski-Harabasz score \citep{chindex}}

In this subsection we present additional results for clustering performance of Algorithm 2 compared to other fair clustering approaches, thus extending the results for the Silhouette scores presented in the main paper for $k=2$. In particular, we consider the Davies-Bouldin index \citep{dbindex} and Calinski-Harabasz score \citep{chindex}. However, these are less indicative and harder to decipher, compared to the more straightforward Silhouette score. This is because these scores are unbounded, and unlike the Silhouette score, do not lie between -1 and 1. It is thus harder to discern and compare different clusterings in an objective manner. The Davies-Bouldin index is an inverse score, in that a lower value signifies better clustering performance, with the lowest possible being 0. This is different from the Calinski-Harabasz score, which gives well-formed clusters higher values, and ill-formed clusters lower ones. Also as in the main paper, we let $k=2$.

We then present the results for Algorithm 2 with the Davies-Bouldin index being used as a clustering performance metric in Table 3. Similar results are shown for the Calinski-Harabasz score in Table 4. It can be observed from the values obtained in the table that we generally have very similar performance to the state-of-the-art fair clustering algorithms, while providing improved fairness (as the results show in the main text). In some cases, we can see that we have better performance, such as in the case of comparisons with the Fair-Lloyd algorithm of \citep{ghadiri2020fair} for Combination \#2 and the \texttt{creditcard} and \texttt{LFW} datasets. We can also observe that while the Sihouette score and Davies-Bouldin index give more reasonable differences in values, values for the Calinski-Harabasz score vary widely and are not easy to justify. However, overall, we can see that Algorithm 2 obtains competitive clustering performance while providing improved fairness on these two metrics as well.

\begin{table}[ht!]
\fontsize{8}{8}\selectfont
\caption{Results for Davies-Bouldin index \citep{dbindex}}
\centering
\begin{tabular}{cccc}
\hline
Clustering-Fairness Combination & Dataset & SOTA Fair Algorithm & Algorithm 3 \\ \hline
\multirow{4}{*}{\begin{tabular}[c]{@{}c@{}}Combination \#1:\\ $\mathcal{C}_{\textrm{k-means}}, \mathcal{F}_{\textrm{balance}}$\end{tabular}} & \texttt{adult} & 1.977075899 & 1.987745353 \\ %\cline{2-4} 
 & \texttt{bank} & 1.32144642 & 1.323176876 \\ %\cline{2-4} 
 & \texttt{creditcard} & 1.535894783 & 1.545304948 \\ %\cline{2-4} 
 & \texttt{LFW} & 1.955060538 & 1.957352428 \\ \hline
\multirow{4}{*}{\begin{tabular}[c]{@{}c@{}}Combination \#2:\\ $\mathcal{C}_{\textrm{k-means}}, \mathcal{F}_{\textrm{social}}$\end{tabular}} & \texttt{adult} & 0.253210096 & 0.253863016 \\ %\cline{2-4} 
 & \texttt{bank} & 1.320149817 & 1.352818868 \\ %\cline{2-4} 
 & \texttt{creditcard} & 1.549144574 & 1.346206321 \\ %\cline{2-4} 
 & \texttt{LFW} & 1.964263126 & 1.6180073 \\ \hline
\multirow{4}{*}{\begin{tabular}[c]{@{}c@{}}Combination \#3:\\ $\mathcal{C}_{\textrm{spectral}}, \mathcal{F}_{\textrm{balance}}$\end{tabular}} & \texttt{adult} & 1.855412037 & 1.939762524 \\ %\cline{2-4} 
 & \texttt{bank} & 1.411452861 & 9.265055739 \\ %\cline{2-4} 
 & \texttt{creditcard} & 1.839992765 & 3.568314803 \\ %\cline{2-4} 
 & \texttt{LFW} & 1.918287593 & 2.04272442 \\ \hline
\end{tabular}
\end{table}

\begin{table}[ht!]
\fontsize{8}{8}\selectfont
\caption{Results for Calinski-Harabasz score \citep{chindex}}
\centering
\begin{tabular}{cccc}
\hline
Clustering-Fairness Combination & Dataset & SOTA Fair Algorithm & Algorithm 3 \\ \hline
\multirow{4}{*}{\begin{tabular}[c]{@{}c@{}}Combination \#1:\\ $\mathcal{C}_{\textrm{k-means}}, \mathcal{F}_{\textrm{balance}}$\end{tabular}} & \texttt{adult} & 1745.35127 & 1670.931497 \\ %\cline{2-4} 
 & \texttt{bank} & 14025.77526 & 13935.82229 \\ %\cline{2-4} 
 & \texttt{creditcard} & 6411.585792 & 6365.048578 \\ %\cline{2-4} 
 & \texttt{LFW} & 3208.171468 & 3200.597262 \\ \hline
\multirow{4}{*}{\begin{tabular}[c]{@{}c@{}}Combination \#2:\\ $\mathcal{C}_{\textrm{k-means}}, \mathcal{F}_{\textrm{social}}$\end{tabular}} & \texttt{adult} & 2170.257982 & 2054.422064 \\ %\cline{2-4} 
 & \texttt{bank} & 13923.43025 & 14677.25017 \\ %\cline{2-4} 
 & \texttt{creditcard} & 6411.04393 & 10905.30975 \\ %\cline{2-4} 
 & \texttt{LFW} & 3193.024907 & 6293.817629 \\ \hline
\multirow{4}{*}{\begin{tabular}[c]{@{}c@{}}Combination \#3:\\ $\mathcal{C}_{\textrm{spectral}}, \mathcal{F}_{\textrm{balance}}$\end{tabular}} & \texttt{adult} & 171.7174597 & 128.9134408 \\ %\cline{2-4} 
 & \texttt{bank} & 284.5544472 & 6.931179858 \\ %\cline{2-4} 
 & \texttt{creditcard} & 197.9853811 & 51.74285991 \\ %\cline{2-4} 
 & \texttt{LFW} & 249.6535979 & 181.497717 \\ \hline
\end{tabular}
\end{table}

%\clearpage
\section{Empirical Running-Time Analysis}
In this subsection we analyze the time complexity of our proposed algorithms. We provide empirical results for all three algorithms on the 4 real-world datasets. As the results show, our algorithms have somewhat higher running times compared to other fair algorithms and vanilla clustering. However, this is precedented as our algorithms provide more generalizablity, and hence incur an increase in time complexity due to the inherent trade-off between the whitebox and blackbox setting. Despite this we do not feel our algorithms are prohibitively expensive in terms of running times, and can provide improved fairness over state-of-the-art fair clustering algorithms. Furthermore, for all experiments that follow, we run each algorithm 10 times, and then present the average. Finally, all experiments are conducted on a machine with an	Intel i7-8750H CPU (2.20GHz and 6 cores).

\textbf{Algorithm 2.} It is not easy to analyze time complexity for Algorithm 2 and compare it fairly to the other state-of-the-art algorithms, as the running time depends heavily on how it is parameterized and the initial size of $V$. For this reason, for comparison, we fix the antidote dataset size to 1 and use the same values for the other parameters mentioned in the main text (such as $k=2$), and try to improve on the fairness cost of the original vanilla clustering. Similarly, for other state-of-the-art fair clustering algorithms, we attempt this same task for the same parameters, and then compare the time it takes for both algorithms (ours and the SOTA) to achieve improved fairness over the vanilla clustering. We do this for all 4 datasets and for all 3 combinations of fairness cost and clustering objectives (as in the main paper). Our running times are tabulated as Table 5. Note that we use the Python implementation provided in \citep{bera2019fair} for comparison for Combination \#1, the MATLAB implementation of the Fair-Lloyd algorithm \citep{ghadiri2020fair} provided by the authors for comparison for Combination \#2, and implement the algorithm of \citep{kleindessner2019guarantees} in Python for Combination \#3.

It can be seen that our algorithms are inherently slower than the fair algorithms, but are not prohibitively slow, and provide improved generalization than other approaches (which are designed specifically for the clustering algorithm and fairness notion in the combination). Also note that running times for all datasets in Combination \#3 appear similar since all datasets were subsampled to 1000, as the fair algorithm of \citep{kleindessner2019guarantees} cannot scale to larger data. This is not the case for the other combinations.

\begin{table}[ht!]
\fontsize{8}{8}\selectfont
\caption{Running-Time Results for Algorithm 2}
\centering
\begin{tabular}{cccc}
\hline
Clustering-Fairness Combination & Dataset & Running Time (Ours) & Running Time (SOTA) \\ \hline
\multirow{4}{*}{\begin{tabular}[c]{@{}c@{}}Combination \#1: \\ $\mathcal{C}_{\textrm{k-means}}, \mathcal{F}_{\textrm{balance}}$\end{tabular}} & \texttt{adult} & 27.3s & 0.359s \\ %\cline{2-4} 
 & \texttt{bank} & 55.234s & 1.71s \\ %\cline{2-4} 
 & \texttt{creditcard} & 104.67s & 1.39s \\ %\cline{2-4} 
 & \texttt{LFW} & 68.5s & 0.797s \\ \hline
\multirow{4}{*}{\begin{tabular}[c]{@{}c@{}}Combination \#2: \\ $\mathcal{C}_{\textrm{k-means}}, \mathcal{F}_{\textrm{social}}$\end{tabular}} & \texttt{adult} & 32.9s & 0.28s \\ %\cline{2-4} 
 & \texttt{bank} & 77.3s & 0.806s \\ %\cline{2-4} 
 & \texttt{creditcard} & 116s & 1.85s \\ %\cline{2-4} 
 & \texttt{LFW} & 120.5s & 14.594s \\ \hline
\multirow{4}{*}{\begin{tabular}[c]{@{}c@{}}Combination \#3: \\ $\mathcal{C}_{\textrm{spectral}}, \mathcal{F}_{\textrm{balance}}$\end{tabular}} & \texttt{adult} & 80s & 0.337s \\ %\cline{2-4} 
 & \texttt{bank} & 75.5s & 0.317s \\ %\cline{2-4} 
 & \texttt{creditcard} & 80.5s & 0.385s \\ %\cline{2-4} 
 & \texttt{LFW} & 84.5s & 0.507s \\ \hline
\end{tabular}
\end{table}

\textbf{Algorithm 1.} Since for Algorithm 1, we are considering a simpler and not practically utilized problem, fair clustering algorithms for the particular combination of social fairness and SON clustering, do not exist. Hence, we compare the running time of Algorithm 1 (solving the KKT reformulation with CVX) with the time it takes CVX to solve the vanilla SON objective. We also let $\lambda = 0.01$ for all experiments. These results are shown in Table 6. As we can see, while the running times are not prohibitive, they are more than vanilla SON, which is to be expected. 

\begin{table}[ht!]
\fontsize{8}{8}\selectfont
\caption{Running-Time Results for Algorithm 1}
\centering
\begin{tabular}{ccc}
\hline
Dataset & Running Time (Ours) & Running Time (Vanilla SON) \\ \hline
\texttt{adult} & 14s & 3.03s \\ %\hline
\texttt{bank} & 3.11s & 1.44s \\ %\hline
\texttt{creditcard} & 127s & 15.6s \\ %\hline
\texttt{LFW} & 117s & 55.2s \\ \hline
\end{tabular}
\end{table}

%\newpage
\section{Code and Implementations}
Throughout the paper, we explicitly specify the libraries, packages, and existing approaches we utilize to implement our algorithms. Furthermore, we implement all our code in Python and aim to open-source it.

\bibliographystyle{unsrtnat.bst}
\bibliography{ref.bib}

%% file: introduction.tex
\looseness-1 With the increasing application of machine learning (ML) algorithms in modern society, the design of fair variants to traditional ML algorithms is an important concern. Vanilla ML algorithms do not account for the biases present in training data against certain \textit{minority protected groups}, and hence, might reinforce them. Furthermore, clustering has been widely used to find meaningful structures, explanatory underlying processes, generative features, and groupings inherent in a set of examples. It plays a significant role in most modern data science applications, such as in medicine \citep{li2011integrating}, vision \citep{lu2006combined}, language modeling \citep{rosa2011topical}, financial decisions \citep{financial}, and various societal resource allocation problems. Thus, ensuring fairness with respect to protected groups is an important issue for clustering algorithms.

\looseness-1 Currently, many different \textit{group-level} notions for fairness in clustering exist, such as \textit{balance} \citep{chierichetti2017fair}, \textit{proportionality} \citep{chen2019proportionally}, \textit{social fairness} \citep{ghadiri2020fair}, among others. Traditionally, to make clustering outputs fair with respect to a specific notion of fairness, fair variants to clustering algorithms need to be proposed. Given that many different clustering algorithms exist, each fair variant proposed requires individual analysis, and possesses different theoretical guarantees. Moreover, if fairness notions or clustering algorithms are changed in a deployed real-world system, the corresponding fair algorithms would also have to be reimplemented. Therefore, instead of coming up with new fair algorithms for each fairness definition and each clustering algorithm, we propose an alternate approach to ensuring fairness for clustering. Inspired by recent research on \textit{adversarial attacks} and \textit{data poisoning}, we aim to augment the dataset with \textit{antidote} data points such that when we use vanilla clustering on this new combined dataset, fairness constraints are met. Thus, instead of changing the clustering algorithm to ensure fairness, we \textit{find} an augmented dataset for which the specified fairness constraints are met when vanilla clustering is undertaken on it. Our approach is therefore applicable in very general case scenarios where group-level fairness on the original dataset can be achieved for any arbitrary choice of center-based clustering algorithm and fairness definition. Note that we aim to make clustering fair in the \textit{pre-clustering} stage as opposed to the \textit{in-clustering} stage, unlike most research on fair clustering.  %Along with the completely general problem setting where no assumptions can be made about the fairness definition or clustering algorithms, we also consider the case when we have convex fairness definitions and clustering algorithms. Another problem setting we consider is when we are given some estimates of \textit{fair target} centers (possibly obtained through some fair clustering algorithms or other domain-specific knowledge), and how we can use this information to solve the bi-level optimization for center-based clustering objectives. For this setting we only consider the following clustering objectives: k-medoids, k-means, and k-median.

\looseness-1 Data augmentation to improve fairness was first proposed by \citep{rastegarpanah2019fighting} for recommendation systems. The authors coined the term \textit{antidote} data for the data points added to the original dataset. However, since recommendation systems and clustering algorithms differ widely, their problem formulation and techniques do not translate to clustering. The antidote data problem for clustering is then as follows: \textit{given a dataset $U$, can we compute (antidote) data $V$ such that when we cluster on $U\cup V$ we obtain a fair clustering output for a chosen fairness notion and clustering algorithm?}

We answer this question in the affirmative by proposing a general bi-level formulation of the antidote data problem for clustering. There are also a number of reasons as to why we cannot reuse existing approaches for adversarial attacks on clustering algorithms, which makes our antidote data formulation (and subsequent algorithmic solutions) novel contributions. Firstly, research on adversarial attacks against clustering is sparse, with only two recent papers since 2018 \citep{chhabra2020suspicion, cina2020}. Secondly, these approaches are defined for specific adversarial objectives, and generally aim to change cluster assignments for points near the clustering decision boundary (Theorem 1 in \citep{chhabra2020suspicion}). However, our bi-level formulation requires the antidote data addition to lead to very specific clustering outcomes that improve fairness irrespective of where points lie in clusters. In summary, we make the following contributions:
\begin{itemize}[wide]
    \item We propose an alternative approach to group-level fair clustering, where we augment the original dataset with data points (\textit{antidote} data) such that when we use vanilla clustering on this new combined dataset, fairness is improved. This is the first work that utilizes data augmentation and antidote points for improving fairness in clustering. In contrast, existing works on fair clustering modify the clustering algorithm specific to a notion of group-level fairness. 
    \item We consider two problem settings for the proposed general bi-level formulation: 1) convex group-level fairness notions and convex center-based clustering objectives, and 2) general group-level fairness notions and general center-based clustering objectives. 
    \item We provide algorithms and analysis for each of these settings, and conduct extensive experiments on real-world datasets for multiple clustering algorithms and fairness notions to demonstrate the efficacy and generality of our approaches. 
    \item We also compare our algorithms to state-of-the-art fair clustering algorithms in terms of fairness, and clustering performance, and find that we achieve improved results on all metrics. 
\end{itemize}

%% file: problem_statement.tex
\subsection{Proposed Problem}

The original dataset is denoted as $U \in \mathbb{R}^{n \times d}$. This is the dataset we wish to augment with some antidote data points such that certain fairness constraints are met when we cluster on the augmented dataset. Furthermore for a matrix $M$, let $M_i$ and $M^i$ denote the $i$-th row and $i$-th column respectively. To start, we first define the clustering problem on $U$. A center-based clustering objective, $\mathcal{C}$, takes in a dataset as input (such as $U$) and outputs a set of $k$ centers $\mu \in \mathbb{R}^{k \times d}$, where $k \leq n$. That is, a clustering objective induces a $k$-partition set of the data, where each sample in the dataset is uniquely mapped to a center $\mu_i \in \mu$ where $\mu \in \mathbb{R}^d$. For example, the k-means clustering objective on $U$ can be defined as $\mathcal{C}_{\textrm{k-means}}(U) \coloneqq \mu = \argmin_{\mu' \in \mathbb{R}^{k \times d}} \sum_{x\in U}\min_{i \in [k]}||x - \mu_i'||^2$. 

We denote the group-level fairness notion as $\mathcal{F}: (\mu, U) \rightarrow \mathbb{R}$. That is, the fairness notion takes as input the set of centers from a clustering algorithm and the original dataset, and outputs a fairness cost. The goal of improving fairness is to then minimize $\mathcal{F}$. It is important to note that fairness will be evaluated only on the original real dataset $U$. Moreover, as we will see, all group-level fairness notions can be defined this way. %as the antidote points are dummy points and do not belong to any protected groups. Moreover, as we will see, all fairness notions can be defined this way. %We will define the \textit{balance} \citep{bera2019fair} (reformulated as a fairness cost) and \textit{social fairness} cost \citep{ghadiri2020fair} notions later in this section as we will be using them throughout the paper.

\textbf{The General Problem.} We now state the antidote data problem for improving fairness. We aim to add a set of data points $V$ to $U$, such that when we cluster on $U\cup V$ and obtain centers $\mu$, $\mathcal{F}(\mu, U)$ is less than some given value $\alpha$. The cost of adding points can be defined as the size of set $V$, and hence, we aim to add as few points as possible. The general bi-level optimization problem is as follows:

\vspace{-0.3cm}
\begin{equation}\tag{P1}\label{p1}
 \begin{aligned}
\min_{V, \, \mu} \quad & |V| \\
\textrm{s.t.} \quad & \mathcal{F}(\mu, U) \leq \alpha \\
& \mu = \mathcal{C}(U \, \textstyle \cup \, V)\\
\end{aligned}   
\end{equation}

\textbf{Relaxation \ref{p1.r}.} In the paper, we also consider a relaxed formulation of problem P1. This relaxation allows us to propose algorithms that in turn also solve problem P1 indirectly. The idea is to fix the size of the antidote dataset $|V| \leq \overline{V}_s$ for a given $\overline{V}_s \in \mathbb{R}$, and optimize the fixed-set $V$ so that we only minimize $\mathcal{F}$ in the upper-level problem. Since minimizing the fairness cost is now the upper-level objective, we can also omit writing it as a constraint using $\alpha$:

\vspace{-0.3cm}
\begin{equation}\tag{P1.R}\label{p1.r}
 \begin{aligned}
\min_{V, \, \mu} \quad & \mathcal{F}(\mu, U) \\
\textrm{s.t.} \quad & \mu = \mathcal{C}(U \, \textstyle \cup \, V)\\
\quad & |V| \leq \overline{V}_s\\
\end{aligned}   
\end{equation}

\subsection{Definitions}

We now define the group-level fairness costs we use in the paper. Consider some $g \in \mathbb{Z}^+$ number of protected groups that comprise $U$. Each protected group has an index $j \in [g]$ and contains a certain number of points of $U$. For simplicity of notation we also assume that a mapping function $\psi(U,j)$ exists which takes in as input $U$ and an integer $j$, where $1 \leq j \leq g$, and gives us the set of points of $U$ which belong to the protected group $j$. Now we can define the \textit{social fairness} cost of Ghadiri et al \citep{ghadiri2020fair}. This was originally proposed for k-means clustering, but it fits well with any center-based clustering objective where Euclidean distance is used as the clustering distance metric. %In our experiments, we ensure that this fairness cost is used only with clustering algorithms for which it is well-defined.

%Now, each of the problem settings for problem P1 either involves (1) making no assumptions about $\mathcal{C}$ and $\mathcal{F}$, (2) assuming both $\mathcal{C}$ and $\mathcal{F}$ are convex functions, or (3) have some estimate of fair target centers $\mu^{\dagger}$ such that $\mathcal{F}(\mu^{\dagger}, U) \leq \alpha$, to ease solving P1. 

\begin{definition}(\textbf{\textit{Social Fairness}} \citep{ghadiri2020fair}). \textit{Let $\Delta(\mu, U) = \sum_{x\in U}\min_{\mu_i \in \mu}||x - \mu_i||^2$ where $U$ is the original dataset and $\mu$ are cluster centers. Then the social fairness cost is defined as:}
\begin{equation*}
    \mathcal{F}_{\textrm{social}}(\mu, U) = \max_{j \in [g]} \biggl \{ \frac{\Delta(\mu, \psi(U,j))}{|\psi(U,j)|} \biggr \}
\end{equation*}
\end{definition}

Next we define the \textit{balance} metric \citep{chierichetti2017fair, bera2019fair}. Traditionally, \textit{balance} is a fairness metric that is not a cost, and is maximized. To fit within our framework, we frame it as a cost by multiplying it with $-1$, and name it the \textit{balance cost}. Again, for simplicity of notation, we assume a mapping function $\phi(U, \mu, i)$ exists which takes in as input $U$, $\mu$, and a cluster label $i \in [k]$ and gives us the points in $U$ which belong to cluster $i$. Note that obtaining cluster labels is trivial as for each $x \in U$ the corresponding label can be obtained as $i = \argmin_{i' \in [k]}||x - \mu_{i'}||$.

\begin{definition}(\textbf{\textit{Balance Cost}} \citep{bera2019fair}). \textit{Let $U$ be the original dataset and $\mu \in \mathbb{R}^{k \times d}$ be the set of cluster centers. Define the following ratio $R(i,j) = \frac{|\psi(U,j)|/|U|}{|\psi(U,j) \cap \phi(U,\mu,i)|/|\phi(U,\mu,i)|}$ which signifies the ratio between the proportion of points of group $j$ in $U$ and proportion of group $j$ points in cluster $i$. The balance cost $\mathcal{F}_{\textrm{balance}} \in [-1,0]$ is then defined:}
\begin{equation*}
    \mathcal{F}_{\textrm{balance}}(\mu, U) = -\min_{i \in [k], j\in [g]}\biggl \{ \min \biggl \{ R(i,j), \frac{1}{R(i,j)} \biggr \} \biggr \}
\end{equation*}
\end{definition}

%% file: proposed_approaches.tex
\looseness-1 We consider problem P1 under 2 different settings and provide algorithms and analysis for each: (1) \textbf{Convex $\mathcal{C}$ and Convex $\mathcal{F}$}, and (2) \textbf{General $\mathcal{C}$ and General $\mathcal{F}$}. While setting (1) comprises more of a toy problem as clustering objectives used in practice are rarely convex, solving problem \ref{p1} for setting (2) is quite challenging. For the first setting with convex functions, we can reduce the bi-level problem to a single-level optimization, allowing us to utilize off-the-shelf solvers to obtain $V$. For the general setting, the antidote data problem is significantly harder and we resort to using zeroth-order optimizers as part of our proposed solution to finding a feasible $V$. 

\subsection{Convex $\mathcal{C}$ and Convex $\mathcal{F}$}
For this setting, we assume that both $\mathcal{C}$ and $\mathcal{F}$ are convex functions. Assuming convexity allows us to effectively reduce the bi-level problem to a single-level form, which can then be provided to off-the-shelf convex/non-convex solvers for optimization. In particular, we exploit the convexity of the functions by replacing the lower-level problem with its Karush-Kuhn-Tucker (KKT) optimality conditions as constraints for the upper-level problem. Since the lower-level clustering problem is convex, the KKT conditions are necessary and sufficient to ensure optimality \citep{dempe2002foundations}. %This is not possible for the general case (problem \ref{p1}), due to the hierarchical nature of the optimization, and hence, we resorted to using derivative-free blackbox methods that do not guarantee feasible solutions. In particular, we exploit the convexity of the functions by replacing the lower-level problem with its Karush-Kuhn-Tucker (KKT) optimality conditions as constraints for the upper-level problem. Since the lower-level clustering problem is convex, the KKT conditions are necessary and sufficient to ensure optimality \citep{dempe2002foundations}. 

As optimizing bi-level problems is in general NP-Hard \citep{sinha2017review}, and problem \ref{p1} contains an NP-Hard cardinality minimization problem \citep{abdi2013cardinality} as the upper-level objective, we use the relaxed form \ref{p1.r} to indirectly solve \ref{p1}. This involves fixing $|V|$ as an input hyperparameter and optimizing $V$ so as to minimize $\mathcal{F}$, without considering $\alpha$. We then use the convexity of the lower-level problem to obtain a single-level reduction from this bi-level problem by replacing the lower-level problem with its KKT constraints. When we minimize this reduced single-level problem, we effectively minimize \ref{p1.r}.

\looseness-1 We describe our approach as Algorithm 1. We aim to solve problem \ref{p1.r} using our algorithm, and in each iteration try to find a suitable $V$ to optimize using the reduced single-level problem (obtained via KKT conditions). In each iteration of the algorithm, we start by fixing the size of $V$ to some $V_s$, and obtain $\mathcal{F}$ after optimizing $V$. If this fairness cost is less than $\alpha$, we can exit, otherwise we increase the size of $V$ (denoted as $V_s$) by $\xi \in \mathbb{Z}^+$ for the next iteration and continue. Algorithm 1 can also exit if the constraint is not met, if a certain number of iterations are exceeded, or if $|V|$ grows to an unacceptable value. We omit these details from Algorithm 1 for simplicity, but they can be easily implemented. %Similar to Algorithm 1, we can exit in the while loop after a certain number of iterations or if $|V| \gg |U|$. 

\looseness-1 Not many widely used convex formulations for clustering algorithms exist except for sum-of-norms (SON) clustering \citep{lindsten2011just, son-hier}, which is strongly convex. SON clustering has been shown to be a convex relaxation to both k-means clustering \citep{lindsten2011just} and hierarchical agglomerative clustering \citep{son-hier}. Below, we analyze SON clustering in the context of Algorithm 1. For the fairness notion, we utilize $\mathcal{F}_{\textrm{social}}$ which is clearly convex and well-defined for SON clustering. We first define the SON clustering objective. It is important to note that we modify the notation-- since the objective is convex, the number of clusters are not discretely defined, but obtained via a regularization parameter $\lambda$. Centers are represented as a $\mathbb{R}^{n \times d}$ matrix as there is no explicitly defined $k$, but note there will only be some unique $k\leq n$ centers decided by the parameterization of $\lambda$. The objective is as follows: $ \mathcal{C}_{\textrm{SON}}(U) \coloneqq \mu = \argmin_{\mu' \in \mathbb{R}^{n \times d}} \frac{1}{2}\sum_{j=1}^n||U_j - \mu_j'||^2 + \lambda\sum_{i<j}||\mu_i' - \mu_j'||$.

%\vspace{-0.4cm}
%\begin{equation*}
%    \mathcal{C}_{\textrm{SON}}(U) \coloneqq \mu = \argmin_{\mu' \in \mathbb{R}^{n \times d}} \frac{1}{2}\sum_{j=1}^n||U_j - \mu_j'||^2 + \lambda\sum_{i<j}||\mu_i' - \mu_j'||
%\end{equation*}

\looseness-1 Let $V_s^{(t)}$ denote the size $V_s$ of $V$ in iteration $t$ of Algorithm 1 (line 2). The number of centers we have will be $\mu \in \mathbb{R}^{m \times d}$ where $m = n + V_s^{(t)}$ for $U \cup V$. To derive the KKT conditions we first reformulate the objective. Consider an ordering of all $(\mu_i, \mu_j)$ pairs where all $i < j$. We can let each of the $m$ centers $\mu_i$ be a node in a graph $G$. The created ordering essentially enumerates the list of edges $E$ for the graph $G$. We denote this ordering as $O$ where we will have $|E| = |O| = m(m-1)/2$. We also denote the node-arc-incidence matrix \citep{node-arc} for $(G, E)$ as $I \in \mathbb{R}^{m \times |O|}$. We can then rewrite the SON objective, define the dual problem to the reformulation, and derive the KKT conditions (details provided in Section A.2 of appendix). Then the single-level reduction for \ref{p1.r} can be written as follows:

%...derive the KKT conditions (details provided in Section A.3 of appendix).

\iffalse
\vspace{-0.3cm}
\begin{equation}\tag{SON.R}\label{sono}
 \begin{aligned}
\min_{\mu \, \in \, \mathbb{R}^{m \times d}, \; \eta \, \in \, \mathbb{R}^{d \times |O|}} \;\; & \frac{1}{2} ||\mu - (U \cup V)||^2 + \lambda \sum_{i \in O}||\eta^i|| \;\;\textrm{ s.t.} \;\;\,\, \mu^{T}I - \eta = 0\\
\end{aligned}   
\end{equation}
\fi

%We can define the dual problem to the \ref{sono} formulation and derive the KKT conditions (details provided in Section A.3 of appendix). Then the single-level reduction for \ref{p1.r} can be written as:

\vspace{-0.3cm}
\begin{equation*}
 \begin{aligned}
\min_{V, \, \mu, \, \eta, \, \theta, \, \zeta } \quad & \mathcal{F}_{\textrm{social}}(\mu, U) \\
\textrm{s.t.} \quad & \theta + \mu - (U\cup V) = 0\\
     & \eta - \max\{0, 1- (1/||\eta + \zeta||)\}(\eta + \zeta) = 0\\
     & \mu^{T}I - \eta = 0\\
     & I\zeta^T - \theta = 0\\
\end{aligned}   
\end{equation*}

Here, $\mu \in \mathbb{R}^{m \times d}, \eta \in \mathbb{R}^{d \times |O|}$ are the primal variables, and $\theta \in \mathbb{R}^{m \times d}$, $\zeta \in \mathbb{R}^{d \times |O|}$ are the dual variables. We also observe that replacing KKT conditions as constraints can introduce non-convexity. All the constraints and objectives are convex, except for one: $\eta - \max\{0, 1-(1/||\eta + \zeta||)\}(\eta + \zeta) = 0$. To approximate this, we can replace it with an affine constraint as $\eta - \gamma(\eta + \zeta) = 0$ where $0 \leq \gamma \leq 1$. Then a convex solver such as CVX \citep{cvxpy} can be used to solve the above problem. Finally, assuming it takes time $T_{\textrm{KKT}}$ to solve the single-level problem, and a feasible antidote dataset $V^*$ exists, Algorithm 1 has a running time of $\mathcal{O}(T_{\textrm{KKT}}|V^*|/\xi)$.

\textbf{Remark.} Since we are solving a convex problem above, the results for this setting are not too difficult to obtain. We thus defer results for Algorithm 1 to the appendix (Section B).

\begin{minipage}{0.48\textwidth}
\begin{algorithm}[H]
\fontsize{8.5}{8.5}\selectfont
%\caption{Proposed Algorithm for Convex $\mathcal{C}$ and $\mathcal{F}$}\label{cvx}
\caption*{\small{\textbf{Algorithm 1}: Convex $\mathcal{C}$ and $\mathcal{F}$}}
\textbf{Input:} $U, \mathcal{C}, \mathcal{F}, V_s, \xi$\\
\textbf{Output:} $V$
 \begin{algorithmic}[1]
 \While{true}
 \State \textbf{initialize} $V$ arbitrarily with $|V| = V_s$
 \State \textbf{reduce} problem \ref{p1.r} by replacing $\mathcal{C}(U\cup V)$ with its KKT conditions as constraints
 \State \textbf{solve} this single-level problem for optimal $V$
 \State \textbf{if} $\mathcal{F}(\mu, U) \leq \alpha$ \textbf{return} $V$ \textbf{else} $V_s \leftarrow V_s + \xi$
 \EndWhile
 \end{algorithmic}
\end{algorithm}
\end{minipage}
\hfill
\begin{minipage}{0.48\textwidth}
\begin{algorithm}[H]
\fontsize{8.5}{8.5}\selectfont
%\caption{Proposed Algorithm for General $\mathcal{C}$ and $\mathcal{F}$}\label{sre}
\caption*{\small{\textbf{Algorithm 2}: General $\mathcal{C}$ and $\mathcal{F}$}}
\textbf{Input:} $U, \mathcal{C}, \mathcal{F}, \mathcal{A}, V_s, n', \xi$\\
\textbf{Output:} $V$
 \begin{algorithmic}[1]
 \While{true}
 \State \textbf{define} $\mu \leftarrow \mathcal{C}(U\cup V)$ and $f(V) \leftarrow \mathcal{F}(\mu, U)$
%\State \textbf{define} $f(V) \leftarrow \mathcal{F}(\mathcal{C}(U\cup V), U)$ and $\mu \leftarrow \mathcal{C}(U\cup V)$
 \State \textbf{initialize} $V$ arbitrarily with $|V| = V_s$
 \State \textbf{optimize} $V$ using \texttt{SRE}($n', f(V), \mathcal{A}$) 
 \State \textbf{obtain} optimized $V$ and $\mathcal{F}(\mu, U)$ from \texttt{SRE} \& $\mathcal{A}$
 \State \textbf{if} $\mathcal{F}(\mu, U) \leq \alpha$ \textbf{return} $V$ \textbf{else} $V_s \leftarrow V_s + \xi$
 \EndWhile
 \end{algorithmic}
\end{algorithm}
\end{minipage}

\subsection{General $\mathcal{C}$ and General $\mathcal{F}$}
\looseness-1 In this setting, we make no assumptions about the clustering objective $\mathcal{C}$ and the fairness cost $\mathcal{F}$. In such a minimal assumption setting where group-level fairness notions as well as center-based clustering objectives can vary widely, it is not trivial to propose algorithms with strong theoretical guarantees. Furthermore, some of the most popular and widely utilized clustering algorithms such as k-means, hierarchical clustering, DBSCAN, etc. possess highly non-convex objectives and are generally optimized via heuristic algorithms (such as Lloyd's algorithm for k-means). In terms of fairness notions for clustering, \textit{balance} is generally the most widely used metric in proposing fair algorithms. As evident in Definition 2.2, it is both non-convex and non-differentiable. 

Furthermore, general bi-level optimization is NP-Hard; even for the simpler case when the upper-level and lower-level problems are linear, a polynomial time algorithm that finds the global optima of the bi-level problem might not exist \citep{sinha2017review}. Since we are dealing with possibly many non-convex upper-level and lower-level problems in this setting, finding a global optima for P1 is not a trivial task. We then resort to finding a locally optimal solution that satisfies our problem constraints. To do this, we relax the NP-Hard upper-level problem which seeks to minimize the size of the antidote dataset $V$. Similar to the convex setting, we are attempting to solve the relaxed formulation \ref{p1.r} (indirectly solving \ref{p1}), where we fix $|V|$ to some given value, and optimize $V$ to minimize $\mathcal{F}$. %If we continue to do this iteratively, we can indirectly solve problem \ref{p1}.

%We then resort to finding a locally optimal solution (if at all) that satisfies our problem constraints. To do this, we relax the upper-level problem which is minimizing the size of the antidote dataset $V$ as it is a cardinality minimization problem, and NP-Hard itself \citep{abdi2013cardinality}. We achieve this by fixing $|V|$ as an input hyper-parameter, and optimizing the fixed-size antidote dataset $V$ to just minimize $\mathcal{F}$ without considering $\alpha$. This is essentially the relaxed formulation of the bi-level optimization P1, which we denoted in the previous section as problem P1.R.

\looseness-1 To solve P1.R, we can use zeroth-order optimization algorithms (such as \texttt{RACOS} \citep{racos}, \texttt{CMAES} \citep{cmaes}, \texttt{IMGPO} \citep{imgpo}). Let such an algorithm be denoted as $\mathcal{A}$. Most zeroth-order optimization algorithms do not scale well with problem input, and hence, cannot usually be applied to data with number of samples $n \geq 1000$ \citep{sre}. However, since our goal is to utilize antidote data on large-scale datasets, the algorithm $\mathcal{A}$ cannot be applied directly to solve P1.R in practice. To circumvent this problem, we propose using the Sequential Random Embedding (\texttt{SRE}) approach of \citep{sre}, which can be used in conjunction with the zeroth-order blackbox optimizer $\mathcal{A}$ to solve P1.R. The \texttt{SRE} approach scales the problem input by projecting it to a low-dimensional setting where it invokes $\mathcal{A}$ to solve the optimization. \texttt{SRE} takes in as input the reduced dimension $n' \ll n$, the objective function $f$ to optimize, and zeroth-order optimization algorithm $\mathcal{A}$. We defer the reader to \citep{sre} for more details on \texttt{SRE}.

\looseness-1 Using the \texttt{SRE} approach, we propose Algorithm 2 for solving \ref{p1.r}. We begin by defining the nested function $f$ to optimize (line 2) which takes in as input some $V$ and outputs the fairness cost $\mathcal{F}(\mu, U)$ where $\mu$ is obtained via $\mathcal{C}(U\cup V)$. The basic idea is to fix $|V|$ to some pre-defined starting value $V_s$ and optimize $V$ using the \texttt{SRE} approach as the back-end (line 3-5). Then, if the constraint $\mathcal{F}(\mu, U) \leq \alpha$ is not met, we increase $|V|$ by some small number $\xi \in \mathbb{Z}^+$ and repeat (line 6). Similar to Algorithm 1, we can exit in the while loop after a certain number of iterations or if $|V| \gg |U|$. %Algorithm 1 can also exit even if the constraint is not met, if a certain number of iterations are exceeded, or if $|V|$ grows to an unacceptable value. We omit these details from Algorithm 1 for simplicity, but they can be easily implemented.

In our experiments for this setting, we use \texttt{RACOS} \citep{racos} as the algorithm $\mathcal{A}$, which is a \textit{Sampling-and-Learning} (SAL) framework. Previous work on SAL approaches allows us to give some weak theoretical results regarding Algorithm 2 on computing a locally optimal solution for Problem \ref{p1.r} and the number of blackbox queries required to do so. We present Theorem \ref{other_paper}, which we have adapted from \citep{yu2014sampling} for our setting. Essentially the result states that the query complexity to compute a locally optimal solution given a fixed-size $V$ to optimize, scales inversely with how effectively $\mathcal{A}$ samples feasible solutions and how many feasible solutions $f$ admits. This does not provide much information from a practical perspective, however through experiments we obtain competitive results on real-world datasets for different combinations of $\mathcal{F}$ and $\mathcal{C}$. Finally, if $\mathcal{A}$ runs for time $T_{\mathcal{A}}$, and assuming a feasible antidote dataset $V^*$ exists, Algorithm 2 has a running time of $\mathcal{O}(T_{\mathcal{A}}|V^*|/\xi)$.

\begin{theorem}\label{other_paper}\citep{yu2014sampling}.
\textit{Let $V^* \in \mathbb{R}^{V_s^{(t)} \times d}$ be a minimizer for the function $f(V)$ in an iteration $t$ of Algorithm 2 and for $\epsilon > 0$ define $X = \{V \in \mathbb{R}^{V_s^{(t)} \times d} \, | \, f(V) - f(V^*) \leq \epsilon\}$. Let $\mathbb{P}_{X}$ denote the average probability of successfully sampling from the uniform distribution over $X$ by algorithm $\mathcal{A}$, and it takes $n_{X}$ samples to realize $\mathbb{P}_{X}$. Then, the number of queries to $f$ that $\mathcal{A}$ makes to compute $\tilde{V}$ s.t. $f(\tilde{V}) - f(V^*) \leq \epsilon$  with probability at least $1 - \delta$ is bounded as $\mathcal{O}(\max\{\frac{\ln{(\delta^{-1})}}{\mathbb{P}_{X}}, n_{X}\})$.}
\end{theorem}

%% file: results.tex
\subsection{Datasets}
We consider four real-world datasets commonly used to evaluate fair clustering algorithms: \texttt{adult} \citep{kohavi1996scaling}, \texttt{bank} \citep{moro2014data}, \texttt{creditcard} \citep{yeh2009comparisons}, and Labeled Faces in the Wild (\texttt{LFW}) \citep{huang2008labeled}. The \texttt{adult} dataset has $10000 \times 5$ samples, and protected groups signify \textit{race} (\textit{white, black, asian-pac-islander, amer-indian-eskimo, other}). The \texttt{bank} dataset has $45211 \times 3$ samples, and protected groups signify \textit{marital status} (\textit{married, single, divorced}). The \texttt{creditcard} dataset has $30000 \times 23$ samples, and the protected groups signify \textit{education} (\textit{higher} and \textit{lower} education). \texttt{LFW} has $13232 \times 80$ samples, and the protected groups signify \textit{sex} (\textit{male, female}). 

We defer the results for Algorithm 1 (with $\mathcal{C}_{\textrm{SON}}$ and $\mathcal{F}_{\textrm{social}}$) to the appendix (Section B) as we are solving a convex problem for which the results can be obtained in a straightforward manner.

\subsection{Results for Algorithm 2}
\looseness-1 We compare Algorithm 2 against vanilla clustering and state-of-the-art fair clustering algorithms. Throughout we let $k=2$ and due to space limitations, present results for $k=3$ and $k=4$ in the appendix (Section C.1). We also compare Algorithm 2 and other fair clustering approaches in terms of clustering performance, using clustering performance metrics such as the Silhouette coefficient \citep{ROUSSEEUW198753}, Calinski-Harabasz score \citep{chindex}, and the Davies-Bouldin index \citep{dbindex}. We use these metrics to unify comparisons across the different clustering algorithms considered in experiments. For all experiments, we choose $\alpha$ to be the fairness cost of the algorithms being compared against (vanilla clustering, fair algorithms) so as to improve on them. We let $\mathcal{A}$ be the \texttt{RACOS} \citep{racos} algorithm, $V_s = 10, n' = 100, \xi = 1$.

\subsubsection{Comparing Algorithm 2 With Vanilla Clustering and Fair Clustering Approaches}
\looseness-1 Since Algorithm 2 can accommodate general $\mathcal{C}$ and $\mathcal{F}$, we experiment on 3 combinations: Combination \#1 with $\mathcal{C}_{\textrm{k-means}}$ and $\mathcal{F}_{\textrm{balance}}$, Combination \#2 with $\mathcal{C}_{\textrm{k-means}}$ and $\mathcal{F}_{\textrm{social}}$, and Combination \#3 where $\mathcal{C}$ is unnormalized spectral clustering, and $\mathcal{F}$ is $\mathcal{F}_{\textrm{balance}}$. The results when comparing against vanilla clustering are shown in Table 1. Vanilla cluster centers are denoted as $\mu^{\textrm{vanilla}}$ and centers obtained via Algorithm 2 are denoted by $\mu$. As can be seen we add very few antidote data points ($|V|/|U|$) and improve on the fairness cost over vanilla clustering. For each of the combination settings considered, we also compare against an equivalent state-of-the-art fair clustering algorithm. For Combination \#1 we consider the algorithm of Bera et al \citep{bera2019fair}, for Combination \#2 we consider the Fair-Lloyd algorithm of Ghadiri et al \citep{ghadiri2020fair}, and for Combination \#3 we consider the algorithm of Kleindessner et al \citep{kleindessner2019guarantees}. Since the approach of \citep{kleindessner2019guarantees} cannot handle large datasets, we subsample each dataset to 1000 samples for Combination \#3. The results are shown in Table 2, and centers obtained from fair clustering algorithms are denoted as $\mu^{\textrm{SOTA}}$. We find that we outperform fair algorithms in terms of lower fairness costs. 

%\vspace{-0.11cm}

% 1 indicating a well-defined clustering and -1 indicating ill-formed clusters
%\vspace{-0.2cm}

%\vspace{-0.2cm}

\begin{table*}[ht!]
\fontsize{8}{8}\selectfont %8
\caption{Comparing fairness costs of Algorithm 2 with vanilla clustering. (Consider Combination \#1 and the \texttt{bank} dataset as an example. The fairness cost for the vanilla cluster centers $\mu^{\textrm{vanilla}}$ is $\mathcal{F}(\mu^{\textrm{vanilla}}, U) = -\textrm{0.3054}$ and $\alpha$ is set to this value to improve on this fairness cost. After Algorithm 2 is run, $V$ is obtained, with size $|V| = $ 0.00011$|U|$. Cluster centers $\mu$ obtained by clustering on $U\cup V$ result in fairness cost $\mathcal{F}(\mu, U) = -\textrm{0.3077}$. This is lower than $\mathcal{F}(\mu^{\textrm{vanilla}}, U)$, leading to improved fairness. Refer to Section 4.2 for more details.)}
\centering
\begin{tabular}{cc||cc||cc}
\hline
Clustering-Fairness Combination & Dataset & $\alpha$ & $|V|/|U|$ & $\mathcal{F}(\mu^{\textrm{vanilla}}, U)$ & $\mathcal{F}(\mu, U)$ \\ \hline%\hline
\multirow{4}{*}{\begin{tabular}[c]{@{}c@{}}Combination \#1:\\ $\mathcal{C}_{\textrm{k-means}}, \mathcal{F}_{\textrm{balance}}$\end{tabular}} & \texttt{adult} & -0.6119 & 0.001 & -0.6119 & \textbf{-0.6196} \\
 & \texttt{bank} & -0.3054 & 0.00011 & -0.3054 & \textbf{-0.3077} \\
 & \texttt{creditcard} & -0.8696 & 0.00017 & -0.8696 & \textbf{-0.8715} \\
 & \texttt{LFW} & -0.8815 & 0.00075 & -0.8815 & \textbf{-0.8821} \\ \hline
\multirow{4}{*}{\begin{tabular}[c]{@{}c@{}}Combination \#2:\\ $\mathcal{C}_{\textrm{k-means}}, \mathcal{F}_{\textrm{social}}$\end{tabular}} & \texttt{adult} & 5.3678 & 0.0005 & 5.3678 & \textbf{4.2104} \\
 & \texttt{bank} & 2.3432 & 0.00022 & 2.3432 & \textbf{2.3416} \\
 & \texttt{creditcard} & 19.740 & 0.00034 & 19.740 & \textbf{19.729} \\
 & \texttt{LFW} & 1406.3411 & 0.00076 & 1406.3411 & \textbf{1406.1676} \\ \hline
\multirow{4}{*}{\begin{tabular}[c]{@{}c@{}}Combination \#3:\\ $\mathcal{C}_{\textrm{spectral}}, \mathcal{F}_{\textrm{balance}}$\end{tabular}} & \texttt{adult} & -0.6458 & 0.001 & -0.6458 & \textbf{-0.6911} \\
 & \texttt{bank} & -0.4811 & 0.00022 & -0.4811 & \textbf{-0.5489} \\
 & \texttt{creditcard} & -0.8384 & 0.00034 & -0.8384 & \textbf{-0.8407} \\
 & \texttt{LFW} & -0.9279 & 0.00076 & -0.9279 & \textbf{-0.9389} \\ \hline
\end{tabular}
\end{table*}

\begin{table*}[ht!]
\fontsize{8}{8}\selectfont %8
\caption{Comparing fairness costs of Algorithm 2 with fair clustering algorithms. (Reads similarly to Table 1.)}%This table reads similarly to Table 2. The cluster centers obtained via the state-of-the-art fair clustering algorithm are $\mu^{\textrm{SOTA}}$ and Algorithm 2 obtains centers $\mu$ with fairness cost $\mathcal{F}(\mu, U) < \mathcal{F}(\mu^{\textrm{SOTA}}, U)$. Refer to Section 4.2 for more details.)}
\centering
\begin{tabular}{cc||cc||cc}
\hline
Clustering-Fairness Combination & Dataset & $\alpha$ & $|V|/|U|$ & $\mathcal{F}(\mu^{\textrm{SOTA}}, U)$ & $\mathcal{F}(\mu, U)$ \\ \hline%\hline
\multirow{4}{*}{\begin{tabular}[c]{@{}c@{}}Combination \#1:\\ $\mathcal{C}_{\textrm{k-means}}, \mathcal{F}_{\textrm{balance}}$\end{tabular}} & \texttt{adult} & -0.6059 & 0.001 & -0.6059 & \textbf{-0.6196} \\
 & \texttt{bank} & -0.3065 & 0.00011 & -0.3065 & \textbf{-0.3077} \\
 & \texttt{creditcard} & -0.8696 & 0.00017 & -0.8696 & \textbf{-0.8715} \\
 & \texttt{LFW} & -0.8816 & 0.00075 & -0.8816 & \textbf{-0.8821} \\ \hline
\multirow{4}{*}{\begin{tabular}[c]{@{}c@{}}Combination \#2:\\ $\mathcal{C}_{\textrm{k-means}}, \mathcal{F}_{\textrm{social}}$\end{tabular}} & \texttt{adult} & 4.2636 & 0.0005 & 4.2636 & \textbf{4.2104} \\
 & \texttt{bank} & 2.3135 & 0.1549 & 2.3135 & \textbf{2.3119} \\
 & \texttt{creditcard} & 18.998 & 0.19 & \textbf{18.998} & \textbf{18.998} \\
 & \texttt{LFW} & 1344.5468 & 0.3999 & 1344.5468 & \textbf{1344.5461} \\ \hline
\multirow{4}{*}{\begin{tabular}[c]{@{}c@{}}Combination \#3:\\ $\mathcal{C}_{\textrm{spectral}}, \mathcal{F}_{\textrm{balance}}$\end{tabular}} & \texttt{adult} & -0.5973 & 0.001 & -0.5973 & \textbf{-0.6911} \\
 & \texttt{bank} & -0.6086 & 0.5 & -0.6086 & \textbf{-0.6899} \\
 & \texttt{creditcard} & -0.8407 & 0.38 & -0.8407 & \textbf{-0.9990} \\
 & \texttt{LFW} & -0.9926 & 0.4 & -0.9926 & \textbf{-0.9997} \\ \hline
\end{tabular}
\end{table*}

\subsubsection{Comparing Clustering Performance}
\looseness-1 For comparison, we use the widely utilized Silhouette score \citep{ROUSSEEUW198753} which lies between $[-1,1]$, with higher scores indicating better clustering performance. We show the results in Figure 1 for each combination setting considered. The fair clusters of Algorithm 2 used here are the same from Table 1. We observe that despite outperforming fair algorithms in terms of fairness, we still exhibit competitive clustering performance. We defer the results for the other performance metrics to the appendix (Section C.2), since those are unbounded and harder to interpret.

 %\vspace{-0.5cm}
% \subsection{Limitations of Proposed Algorithms}
 
  %\begin{wrapfigure}{R}{.5\textwidth}
\begin{figure*}[ht!]
\centering
\begin{subfigure}{.33\textwidth} %.32
  \centering
    \includegraphics[scale=0.1]{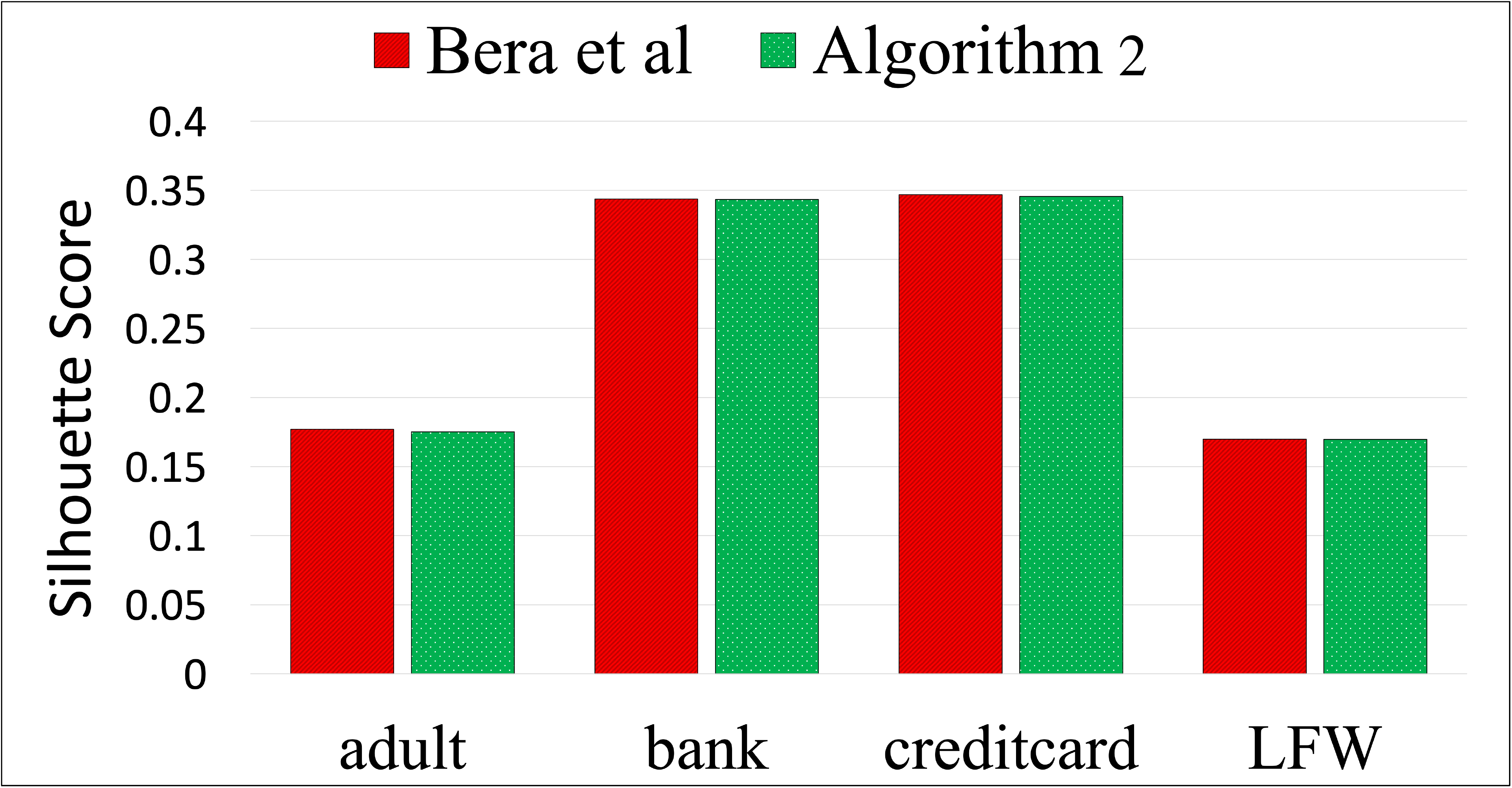}%0.09
  \caption{Combination \#1}
\smallskip
\end{subfigure}\hfill
\begin{subfigure}{.33\textwidth}
  \centering
    \includegraphics[scale=0.1]{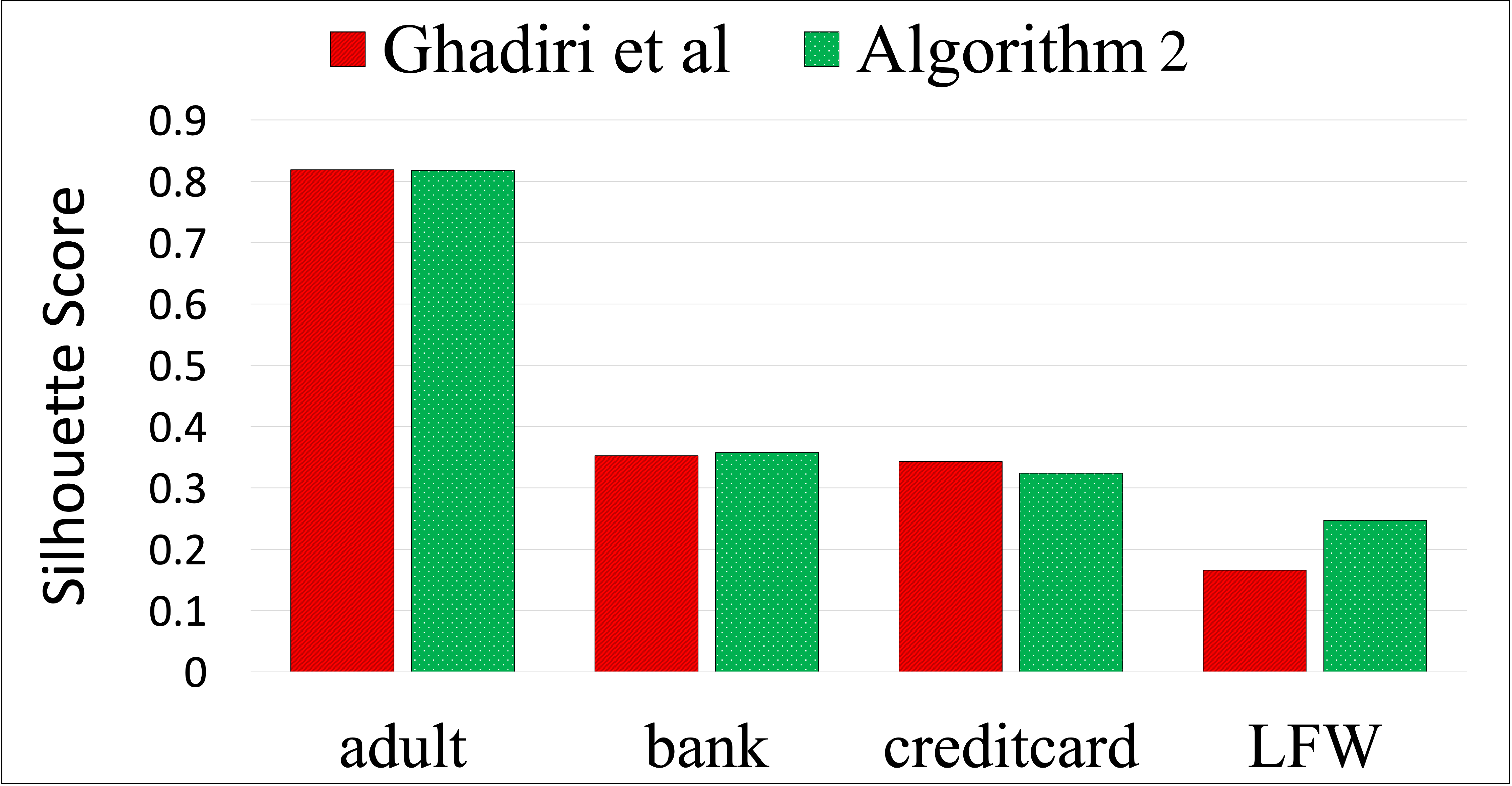}%
\caption{Combination \#2}
\smallskip
\end{subfigure}\hfill
\begin{subfigure}{.33\textwidth}
  \centering
    \includegraphics[scale=0.1]{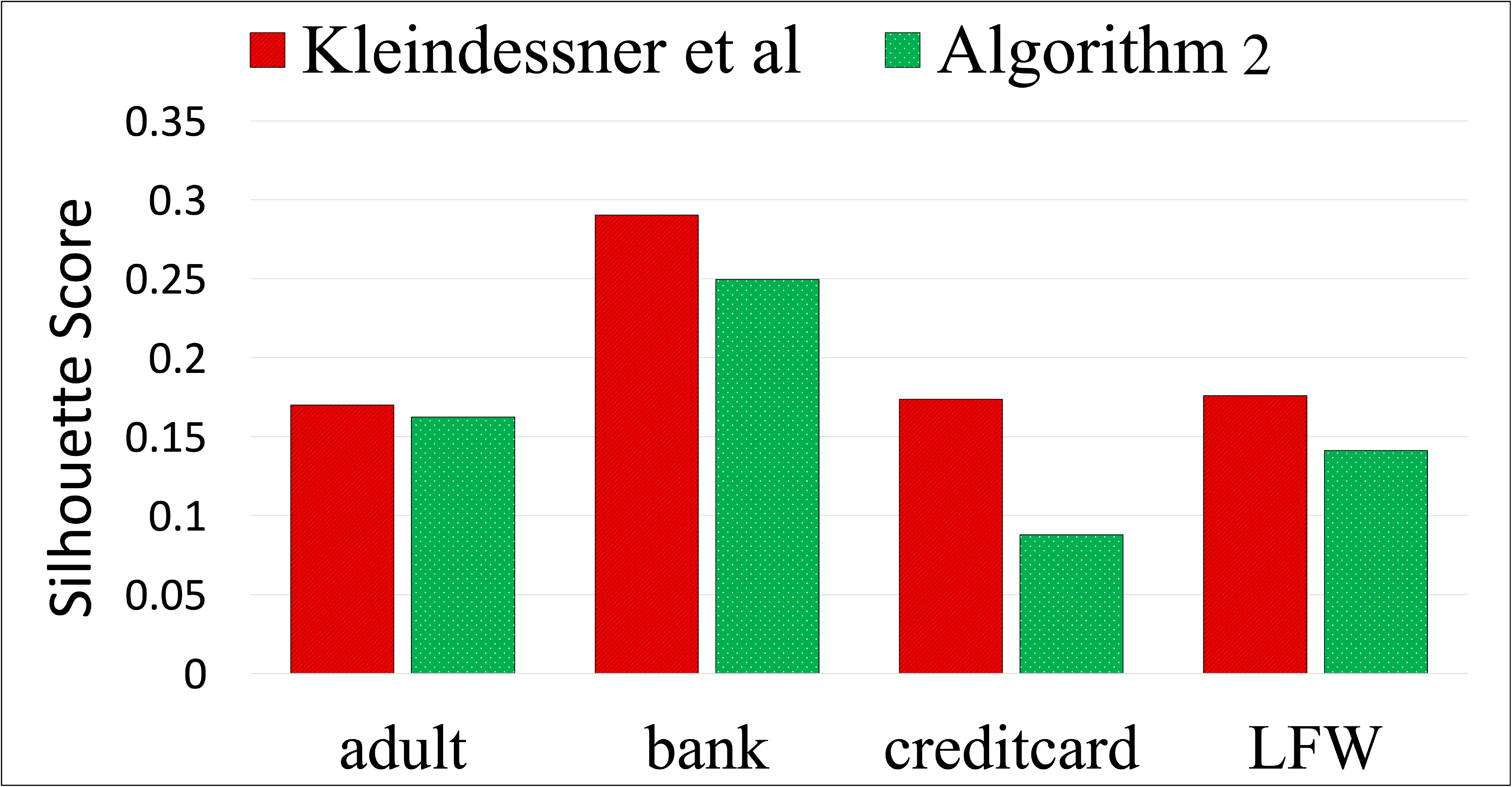}%0.1 scale
  \caption{Combination \#3}
\smallskip
\end{subfigure}
\caption{Comparing clustering performance of Algorithm 2 with fair clustering algorithms using Silhouette scores. (Higher scores indicate better clustering performance. As can be observed, fair clusters obtained via Algorithm 2 achieve similar clustering performance to SOTA algorithms, while providing improved fairness.)}
%\end{wrapfigure}
\end{figure*}

%% file: related_works.tex
\looseness-1 \textbf{Fairness in Machine Learning.} ML algorithms can be made \textit{fair} in three stages of the learning pipeline \citep{caton2020fairness, mehrabi2019survey}-- \textit{before-training} (pre-processing the dataset), \textit{during-training} (changing the ML algorithm), or \textit{after-training} (post-processing the learnt model). Most research on fair clustering focuses on the \textit{during-training} phase \citep{bera2019fair, bercea2018cost, backurs2019scalable, chierichetti2017fair, schmidt2018fair, ziko2019clustering, kleindessner2019guarantees, ghadiri2020fair} and proposes fair clustering algorithms. In their paper, \citep{davidsonmaking} study the \textit{after-training} phase for improving fairness post-clustering. The approaches proposed in our paper are novel since they improve fairness for clustering models in the \textit{before-training} stage. Further, our approaches can accommodate general fairness notions and clustering algorithms. %Another consideration is the fairness notion being employed which varies depending on application context and type of ML model used. In contrast, our approaches can accommodate general fairness notions and clustering algorithms.

\textbf{Machine Teaching.} Our approach in this paper is inspired by the techniques in \textit{machine teaching} literature~\citep{goldman1995complexity,doliwa2014recursive,DBLP:conf/icml/SinglaBBKK14,zhu2015machine,zhu2018overview,DBLP:conf/nips/ChenSAPY18,DBLP:conf/nips/PeltolaCDK19}. Machine teaching studies the interaction between a teacher and a learner where the teacher selects training examples for the learner to learn a specific task. A machine teaching problem can be cast in a bi-level form where the upper-level problem defines the teacher's cost and the lower-level problem defines the learner's method. Variations of this bi-level form can be used to formulate teacher's optimization problem in a variety of learning settings, including supervised learning~\citep{zhu2013machine,liu2016teaching,DBLP:conf/nips/Mansouri0VZS19,DBLP:conf/ijcai/DevidzeMH0S20}, imitation learning~\citep{cakmak2012algorithmic,DBLP:conf/nips/HaugTS18,DBLP:conf/ijcai/KamalarubanDCS19,brown2019machine,DBLP:conf/nips/TschiatschekGHD19}, and reinforcement learning~\citep{DBLP:conf/aaai/ZhangP08,DBLP:conf/sigecom/ZhangPC09,DBLP:conf/icml/RakhshaRD0S20,DBLP:conf/nips/MaZSZ19,DBLP:journals/corr/abs-2011-10824}. In the proposed antidote data problem for clustering, the upper-level problem (teacher's cost) is the cost of adding antidote data, and the lower-level problem (learner) is the clustering algorithm. 

%In its basic form, the formulation of machine teaching focusses The teacher aims to \textit{teach} a specific model to the student, while minimizing the teaching cost. 
%

%Our work also has connections to \textit{machine teaching} \citep{zhu2018overview}, where problems are formulated in a bi-level form. The upper-level problem defines the teacher's cost, and the lower-level problem is the learner. The teacher aims to \textit{teach} a specific model to the student, while minimizing the teaching cost. Machine teaching has been extensively studied for supervised learning \citep{kumar2020average, kumar2021teaching} and reinforcement learning \citep{rakhsha2020policy}, \citep{NEURIPS2019_3de568f8}. In the proposed antidote data problem for clustering, the upper-level problem (teacher's cost) is the cost of adding antidote data, and the lower-level problem (learner) is the clustering algorithm. 

\looseness-1 \textbf{Bi-level Optimization.} Bi-level problems involve a two-level hierarchical optimization. For these, a lower-level problem exists, which influences the solutions for an upper-level problem. Both bi-level optimization and verifying the optimality of an obtained solution are NP-Hard \citep{bilevelnp, bilevelnp2}. This makes finding optimal solutions and evaluating them non-trivial tasks. In the paper, the main problem considered is a complex bi-level optimization, where both upper-level and lower-level problems can be non-convex optimization problems. Many techniques for bi-level programming exist, but most of these assume simple forms for the upper/lower problems, or use evolutionary methods for which theoretical results are hard to provide \citep{sinha2017review}. Despite these challenges, we provide algorithms that obtain feasible solutions to the bi-level problem and outperform state-of-the-art fair clustering approaches.

%\fi

%% file: conclusion.tex
\looseness-1 We propose the antidote data problem for improving group-level fairness in center-based clustering. We provide a more general alternative to traditional approaches aimed at making clustering fair. Instead of proposing new fair variants to clustering algorithms, we augment the original dataset with new \textit{antidote} data points. When regular clustering is undertaken on this new dataset, the clustering output is \textit{fair}. This approach insipred by research on data poisoning attacks, voids the need to come up with new fair algorithms or individual analysis, for different group-level fairness notions or center-based clustering algorithms. Our approach also does not require reimplementation for deployment in actual systems, if the fairness notion or clustering algorithm is changed. We find that our algorithms only need to add a small percentage of points to achieve the given fairness constraints on many real-world datasets without loss of clustering performance. %For future work we aim to provide improved algorithms for the general bi-level problem.

%, making them viable alternatives to traditional fair variants to clustering algorithms.

\looseness-1 A major limitation of our work is running time. We present empirical results in the appendix (Section D). While not prohibitively slow, in comparison to fair clustering algorithms, Algorithm 2 is generally slower and requires careful parameterization for convergence. Similar limitations hold for the other algorithm. However, we believe that despite these shortcomings, our paper opens up an important alternative direction for future research in fair clustering, as our experiments also demonstrate. For future work, we aim to provide faster and more general algorithms for the bi-level problem.

%% file: main.bbl
\begin{thebibliography}{10}

\bibitem{li2011integrating}
Bing~Nan Li, Chee~Kong Chui, Stephen Chang, and Sim~Heng Ong.
\newblock Integrating {S}patial {F}uzzy {C}lustering with {L}evel {S}et
  {M}ethods for {A}utomated {M}edical {I}mage {S}egmentation.
\newblock {\em Computers in biology and medicine}, 41(1):1--10, 2011.

\bibitem{lu2006combined}
Le~Lu and Ren{\'{e}} Vidal.
\newblock Combined {C}entral and {S}ubspace {C}lustering for {C}omputer
  {V}ision {A}pplications.
\newblock In {\em ICML}, volume 148, pages 593--600, 2006.

\bibitem{rosa2011topical}
Kevin~Dela Rosa, Rushin Shah, Bo~Lin, Anatole Gershman, and Robert Frederking.
\newblock Topical {C}lustering of {T}weets.
\newblock {\em Proceedings of the ACM SIGIR: SWSM}, 63, 2011.

\bibitem{financial}
Vijay Hanagandi, Amitava Dhar, and Kevin Buescher.
\newblock Density-based {C}lustering and {R}adial {B}asis {F}unction {M}odeling
  to {G}enerate {C}redit {C}ard {F}raud {S}cores.
\newblock In {\em Proceedings of the {IEEE/IAFE} Conference on Computational
  Intelligence for Financial Engineering}, pages 247--251, 1996.

\bibitem{chierichetti2017fair}
Flavio Chierichetti, Ravi Kumar, Silvio Lattanzi, and Sergei Vassilvitskii.
\newblock Fair {C}lustering {T}hrough {F}airlets.
\newblock In {\em NeurIPS}, pages 5029--5037, 2017.

\bibitem{chen2019proportionally}
Xingyu Chen, Brandon Fain, Liang Lyu, and Kamesh Munagala.
\newblock Proportionally {F}air {C}lustering.
\newblock In {\em ICML}, volume~97, pages 1032--1041, 2019.

\bibitem{ghadiri2020fair}
Mehrdad Ghadiri, Samira Samadi, and Santosh~S. Vempala.
\newblock Socially {F}air k-means {C}lustering.
\newblock In {\em FAccT '21: 2021 {ACM} Conference on Fairness, Accountability,
  and Transparency}, pages 438--448, 2021.

\bibitem{rastegarpanah2019fighting}
Bashir Rastegarpanah, Krishna~P. Gummadi, and Mark Crovella.
\newblock Fighting {F}ire with {F}ire: {U}sing {A}ntidote {D}ata to {I}mprove
  {P}olarization and {F}airness of {R}ecommender {S}ystems.
\newblock In {\em {WSDM}}, pages 231--239, 2019.

\bibitem{chhabra2020suspicion}
Anshuman Chhabra, Abhishek Roy, and Prasant Mohapatra.
\newblock Suspicion-{F}ree {A}dversarial {A}ttacks on {C}lustering
  {A}lgorithms.
\newblock In {\em AAAI}, pages 3625--3632, 2020.

\bibitem{cina2020}
Antonio~Emanuele Cin{\`{a}}, Alessandro Torcinovich, and Marcello Pelillo.
\newblock A {B}lack-box {A}dversarial {A}ttack for {P}oisoning {C}lustering.
\newblock {\em CoRR}, abs/2009.05474, 2020.

\bibitem{bera2019fair}
Suman~Kalyan Bera, Deeparnab Chakrabarty, Nicolas Flores, and Maryam Negahbani.
\newblock Fair {A}lgorithms for {C}lustering.
\newblock In {\em NeurIPS}, pages 4955--4966, 2019.

\bibitem{dempe2002foundations}
Stephan Dempe.
\newblock {\em Foundations of {B}ilevel {P}rogramming}.
\newblock Springer Science \& Business Media, 2002.

\bibitem{sinha2017review}
Ankur Sinha, Pekka Malo, and Kalyanmoy Deb.
\newblock A {R}eview on {B}ilevel {O}ptimization: {F}rom {C}lassical to
  {E}volutionary {A}pproaches and {A}pplications.
\newblock {\em {IEEE} Trans. Evol. Comput.}, 22(2):276--295, 2018.

\bibitem{abdi2013cardinality}
Mohammad~Javad Abdi.
\newblock {\em Cardinality {O}ptimization {P}roblems}.
\newblock PhD thesis, University of Birmingham, 2013.

\bibitem{lindsten2011just}
Fredrik Lindsten, Henrik Ohlsson, and Lennart Ljung.
\newblock {\em Just {R}elax and {C}ome {C}lustering!: A {C}onvexification of
  k-means {C}lustering}.
\newblock Link{\"o}ping University Electronic Press, 2011.

\bibitem{son-hier}
Toby Hocking, Jean{-}Philippe Vert, Francis~R. Bach, and Armand Joulin.
\newblock Clusterpath: an {A}lgorithm for {C}lustering using {C}onvex {F}usion
  {P}enalties.
\newblock In {\em ICML}, pages 745--752, 2011.

\bibitem{node-arc}
André~A. Keller.
\newblock Chapter 3 - elements of technical background.
\newblock In {\em Mathematical Optimization Terminology}, pages 239--298. 2018.

\bibitem{cvxpy}
Steven Diamond and Stephen~P. Boyd.
\newblock {CVXPY:} {A} {P}ython-{E}mbedded {M}odeling {L}anguage for {C}onvex
  {O}ptimization.
\newblock {\em JMLR}, 17:83:1--83:5, 2016.

\bibitem{racos}
Yang Yu, Hong Qian, and Yi{-}Qi Hu.
\newblock Derivative-{F}ree {O}ptimization via {C}lassification.
\newblock In {\em AAAI}, pages 2286--2292, 2016.

\bibitem{cmaes}
Nikolaus Hansen, Sibylle~D. M{\"{u}}ller, and Petros Koumoutsakos.
\newblock Reducing the {T}ime {C}omplexity of the {D}erandomized {E}volution
  {S}trategy with {C}ovariance {M}atrix {A}daptation {(CMA-ES)}.
\newblock {\em Evol. Comput.}, 11(1):1--18, 2003.

\bibitem{imgpo}
Kenji Kawaguchi, Leslie~Pack Kaelbling, and Tom{\'{a}}s Lozano{-}P{\'{e}}rez.
\newblock Bayesian {O}ptimization with {E}xponential {C}onvergence.
\newblock In {\em NeurIPS}, pages 2809--2817, 2015.

\bibitem{sre}
Hong Qian, Yi{-}Qi Hu, and Yang Yu.
\newblock Derivative-{F}ree {O}ptimization of {H}igh-{D}imensional
  {N}on-{C}onvex {F}unctions by {S}equential {R}andom {E}mbeddings.
\newblock In {\em IJCAI}, pages 1946--1952, 2016.

\bibitem{yu2014sampling}
Yang Yu and Hong Qian.
\newblock The {S}ampling-and-learning {F}ramework: {A} {S}tatistical {V}iew of
  {E}volutionary {A}lgorithms.
\newblock In {\em Proceedings of the {IEEE} Congress on Evolutionary
  Computation, {CEC}}, pages 149--158, 2014.

\bibitem{kohavi1996scaling}
Ron Kohavi.
\newblock Scaling {U}p the {A}ccuracy of {N}aive-{B}ayes {C}lassifiers: {A}
  {D}ecision-{T}ree {H}ybrid.
\newblock In {\em KDD}, pages 202--207, 1996.

\bibitem{moro2014data}
S{\'{e}}rgio Moro, Paulo Cortez, and Paulo Rita.
\newblock A {D}ata-driven {A}pproach to {P}redict the {S}uccess of {B}ank
  {T}elemarketing.
\newblock {\em Decis. Support Syst.}, 62:22--31, 2014.

\bibitem{yeh2009comparisons}
I{-}Cheng Yeh and Che{-}hui Lien.
\newblock The {C}omparisons of {D}ata {M}ining {T}echniques for the
  {P}redictive {A}ccuracy of {P}robability of {D}efault of {C}redit {C}ard
  {C}lients.
\newblock {\em Expert Syst. Appl.}, 36(2):2473--2480, 2009.

\bibitem{huang2008labeled}
Gary~B Huang, Marwan Mattar, Tamara Berg, and Eric Learned-Miller.
\newblock Labeled {F}aces in the {W}ild: A {D}atabase {F}or {S}tudying {F}ace
  {R}ecognition in {U}nconstrained {E}nvironments, 2008.

\bibitem{ROUSSEEUW198753}
Peter~J. Rousseeuw.
\newblock Silhouettes: {A} {G}raphical {A}id to the {I}nterpretation and
  {V}alidation of {C}luster {A}nalysis.
\newblock {\em Journal of Computational and Applied Mathematics}, 20:53--65,
  1987.

\bibitem{chindex}
T.~Caliński and J~Harabasz.
\newblock A {D}endrite {M}ethod for {C}luster {A}nalysis.
\newblock {\em Communications in Statistics}, 3(1):1--27, 1974.

\bibitem{dbindex}
David~L. Davies and Donald~W. Bouldin.
\newblock A {C}luster {S}eparation {M}easure.
\newblock {\em {IEEE} Trans. Pattern Anal. Mach. Intell.}, 1(2):224--227, 1979.

\bibitem{kleindessner2019guarantees}
Matth{\"{a}}us Kleindessner, Samira Samadi, Pranjal Awasthi, and Jamie
  Morgenstern.
\newblock Guarantees for {S}pectral {C}lustering with {F}airness {C}onstraints.
\newblock In {\em ICML}, volume~97, pages 3458--3467, 2019.

\bibitem{caton2020fairness}
Simon Caton and Christian Haas.
\newblock Fairness in {M}achine {L}earning: {A} {S}urvey.
\newblock {\em CoRR}, abs/2010.04053, 2020.

\bibitem{mehrabi2019survey}
Ninareh Mehrabi, Fred Morstatter, Nripsuta Saxena, Kristina Lerman, and Aram
  Galstyan.
\newblock A {S}urvey on {B}ias and {F}airness in {M}achine {L}earning.
\newblock {\em CoRR}, abs/1908.09635, 2019.

\bibitem{bercea2018cost}
Ioana~Oriana Bercea, Martin Gro{\ss}, Samir Khuller, Aounon Kumar, Clemens
  R{\"{o}}sner, Daniel~R. Schmidt, and Melanie Schmidt.
\newblock On the {C}ost of {E}ssentially {F}air {C}lusterings.
\newblock In {\em {APPROX/RANDOM}}, volume 145 of {\em LIPIcs}, pages
  18:1--18:22. Schloss Dagstuhl - Leibniz-Zentrum f{\"{u}}r Informatik, 2019.

\bibitem{backurs2019scalable}
Arturs Backurs, Piotr Indyk, Krzysztof Onak, Baruch Schieber, Ali Vakilian, and
  Tal Wagner.
\newblock Scalable {F}air {C}lustering.
\newblock In {\em ICML}, volume~97, pages 405--413, 2019.

\bibitem{schmidt2018fair}
Melanie Schmidt, Chris Schwiegelshohn, and Christian Sohler.
\newblock Fair {C}oresets and {S}treaming {A}lgorithms for {F}air k-means.
\newblock In {\em Approximation and Online Algorithms - 17th International
  Workshop, {WAOA}}, volume 11926 of {\em Lecture Notes in Computer Science},
  pages 232--251, 2019.

\bibitem{ziko2019clustering}
Imtiaz~Masud Ziko, Eric Granger, Jing Yuan, and Ismail~Ben Ayed.
\newblock Clustering with {F}airness {C}onstraints: {A} {F}lexible and
  {S}calable {A}pproach.
\newblock {\em CoRR}, abs/1906.08207, 2019.

\bibitem{davidsonmaking}
Ian Davidson and S.~S. Ravi.
\newblock Making {E}xisting {C}lusterings {F}airer: {A}lgorithms, {C}omplexity
  {R}esults and {I}nsights.
\newblock In {\em AAAI}, pages 3733--3740, 2020.

\bibitem{goldman1995complexity}
Sally~A. Goldman and Michael~J. Kearns.
\newblock On the {C}omplexity of {T}eaching.
\newblock {\em Journal of Computer and System Sciences}, 50(1):20--31, 1995.

\bibitem{doliwa2014recursive}
Thorsten Doliwa, Gaojian Fan, Hans~Ulrich Simon, and Sandra Zilles.
\newblock Recursive {T}eaching {D}imension, {VC}-{D}imension and {S}ample
  {C}ompression.
\newblock {\em JMLR}, 15(1):3107--3131, 2014.

\bibitem{DBLP:conf/icml/SinglaBBKK14}
Adish Singla, Ilija Bogunovic, G{\'{a}}bor Bart{\'{o}}k, Amin Karbasi, and
  Andreas Krause.
\newblock Near-{O}ptimally {T}eaching the {C}rowd to {C}lassify.
\newblock In {\em ICML}, volume~32, pages 154--162, 2014.

\bibitem{zhu2015machine}
Xiaojin Zhu.
\newblock Machine {T}eaching: {A}n {I}nverse {P}roblem to {M}achine {L}earning
  and an {A}pproach {T}oward {O}ptimal {E}ducation.
\newblock In {\em AAAI}, pages 4083--4087, 2015.

\bibitem{zhu2018overview}
Xiaojin Zhu, Adish Singla, Sandra Zilles, and Anna~N. Rafferty.
\newblock An {O}verview of {M}achine {T}eaching.
\newblock {\em CoRR}, abs/1801.05927, 2018.

\bibitem{DBLP:conf/nips/ChenSAPY18}
Yuxin Chen, Adish Singla, Oisin~Mac Aodha, Pietro Perona, and Yisong Yue.
\newblock Understanding the {R}ole of {A}daptivity in {M}achine {T}eaching:
  {T}he {C}ase of {V}ersion {S}pace {L}earners.
\newblock In {\em NeurIPS}, 2018.

\bibitem{DBLP:conf/nips/PeltolaCDK19}
Tomi Peltola, Mustafa~Mert {\c{C}}elikok, Pedram Daee, and Samuel Kaski.
\newblock Machine {T}eaching of {A}ctive {S}equential {L}earners.
\newblock In {\em NeurIPS}, 2019.

\bibitem{zhu2013machine}
Xiaojin Zhu.
\newblock Machine {T}eaching for {B}ayesian {L}earners in the {E}xponential
  {F}amily.
\newblock In {\em NeurIPS}, pages 1905--1913, 2013.

\bibitem{liu2016teaching}
Ji~Liu and Xiaojin Zhu.
\newblock The {T}eaching {D}imension of {L}inear {L}earners.
\newblock {\em JMLR}, 17(162):1--25, 2016.

\bibitem{DBLP:conf/nips/Mansouri0VZS19}
Farnam Mansouri, Yuxin Chen, Ara Vartanian, Xiaojin Zhu, and Adish Singla.
\newblock Preference-based {B}atch and {S}equential {T}eaching: {T}owards a
  {U}nified {V}iew of {M}odels.
\newblock In {\em NeurIPS}, pages 9195--9205, 2019.

\bibitem{DBLP:conf/ijcai/DevidzeMH0S20}
Rati Devidze, Farnam Mansouri, Luis Haug, Yuxin Chen, and Adish Singla.
\newblock Understanding the {P}ower and {L}imitations of {T}eaching with
  {I}mperfect {K}nowledge.
\newblock In {\em IJCAI}, pages 2647--2654.

\bibitem{cakmak2012algorithmic}
Maya Cakmak and Manuel Lopes.
\newblock Algorithmic and {H}uman {T}eaching of {S}equential {D}ecision
  {T}asks.
\newblock In {\em AAAI}, volume~26, 2012.

\bibitem{DBLP:conf/nips/HaugTS18}
Luis Haug, Sebastian Tschiatschek, and Adish Singla.
\newblock Teaching {I}nverse {R}einforcement {L}earners via {F}eatures and
  {D}emonstrations.
\newblock In {\em NeurIPS}, pages 8473--8482, 2018.

\bibitem{DBLP:conf/ijcai/KamalarubanDCS19}
Parameswaran Kamalaruban, Rati Devidze, Volkan Cevher, and Adish Singla.
\newblock Interactive {T}eaching {A}lgorithms for {I}nverse {R}einforcement
  {L}earning.
\newblock In {\em IJCAI}, pages 2692--2700, 2019.

\bibitem{brown2019machine}
Daniel~S Brown and Scott Niekum.
\newblock Machine {T}eaching for {I}nverse {R}einforcement {L}earning:
  {A}lgorithms and {A}pplications.
\newblock In {\em AAAI}, 2019.

\bibitem{DBLP:conf/nips/TschiatschekGHD19}
Sebastian Tschiatschek, Ahana Ghosh, Luis Haug, Rati Devidze, and Adish Singla.
\newblock Learner-aware {T}eaching: {I}nverse {R}einforcement {L}earning with
  {P}references and {C}onstraints.
\newblock In {\em NeurIPS}, 2019.

\bibitem{DBLP:conf/aaai/ZhangP08}
Haoqi Zhang and David~C. Parkes.
\newblock Value-{B}ased {P}olicy {T}eaching with {A}ctive {I}ndirect
  {E}licitation.
\newblock In {\em AAAI}, pages 208--214, 2008.

\bibitem{DBLP:conf/sigecom/ZhangPC09}
Haoqi Zhang, David~C. Parkes, and Yiling Chen.
\newblock Policy {T}eaching through {R}eward {F}unction {L}earning.
\newblock In {\em EC}, pages 295--304, 2009.

\bibitem{DBLP:conf/icml/RakhshaRD0S20}
Amin Rakhsha, Goran Radanovic, Rati Devidze, Xiaojin Zhu, and Adish Singla.
\newblock Policy {T}eaching via {E}nvironment {P}oisoning: {T}raining-time
  {A}dversarial {A}ttacks against {R}einforcement {L}earning.
\newblock In {\em ICML}, volume 119, pages 7974--7984, 2020.

\bibitem{DBLP:conf/nips/MaZSZ19}
Yuzhe Ma, Xuezhou Zhang, Wen Sun, and Jerry Zhu.
\newblock Policy {P}oisoning in {B}atch {R}einforcement {L}earning and
  {C}ontrol.
\newblock In {\em NeurIPS}, pages 14543--14553, 2019.

\bibitem{DBLP:journals/corr/abs-2011-10824}
Amin Rakhsha, Goran Radanovic, Rati Devidze, Xiaojin Zhu, and Adish Singla.
\newblock Policy {T}eaching in {R}einforcement {L}earning via {E}nvironment
  {P}oisoning {A}ttacks.
\newblock {\em CoRR}, abs/2011.10824, 2020.

\bibitem{bilevelnp}
Pierre Hansen, Brigitte Jaumard, and Gilles Savard.
\newblock New {B}ranch-and-{B}ound {R}ules for {L}inear {B}ilevel
  {P}rogramming.
\newblock {\em {SIAM} J. Sci. Comput.}, 13(5):1194--1217, 1992.

\bibitem{bilevelnp2}
L.~Vicente, G.~Savard, and J.~J\'{u}dice.
\newblock Descent {A}pproaches for {Q}uadratic {B}ilevel {P}rogramming.
\newblock {\em J. Optim. Theory Appl.}, 81(2):379–399, May 1994.

\end{thebibliography}
